\documentclass{article}

\usepackage[preprint]{neurips_2025}

\usepackage[utf8]{inputenc} % allow utf-8 input
\usepackage[T1]{fontenc}    % use 8-bit T1 fonts
\usepackage{hyperref}       % hyperlinks
\usepackage{url}            % simple URL typesetting
\usepackage{booktabs}       % professional-quality tables
\usepackage{amsfonts}       % blackboard math symbols
\usepackage{nicefrac}       % compact symbols for 1/2, etc.
\usepackage{microtype}      % microtypography
\usepackage{xcolor}         % colors
\usepackage{wrapfig}
\usepackage{amsthm}
\usepackage{amssymb}
\usepackage{amsmath}
\usepackage{enumitem}
\usepackage{wrapfig}
\usepackage{subfigure}
\usepackage{graphicx}
\usepackage[toc,page,header]{appendix}
\usepackage{minitoc}

\graphicspath{ {./dropbox_article/} }

\theoremstyle{plain}
\newtheorem{thm}{\protect\theoremname}
\theoremstyle{definition}
\newtheorem{defn}[thm]{\protect\definitionname}
\theoremstyle{plain}

\theoremstyle{remark}
\newtheorem{rem}[thm]{\protect\remarkname}

\theoremstyle{plain}
\newtheorem{assumption}{\textbf{Assumption}}

\theoremstyle{plain}
\newtheorem{proposition}{\textbf{Proposition}}

\makeatother

\usepackage{babel}
\providecommand{\definitionname}{Definition}
\providecommand{\lemmaname}{Lemma}
\providecommand{\remarkname}{Remark}
\providecommand{\theoremname}{Theorem}

\title{Kernel-Smoothed Scores for Denoising Diffusion: A Bias-Variance Study}

\author{%
  Franck Gabriel \thanks{These authors contributed equally to this work.} \\
  Université Claude Bernard Lyon 1 \\
  Laboratoire SAF, ISFA, France\\
  \texttt{franck.gabriel@univ-lyon1.fr} \\
  \And
  Francois G. Ged\footnotemark[1] \\
  Department of Mathematics \\
  University of Vienna, Austria \\
  \texttt{fged.math@gmail.com} \\
  \AND
  Maria Han Veiga \\
  Department of Mathematics  \\
  The Ohio State University, USA \\
  \texttt{hanveiga.1@osu.edu} \\
  \And
  Emmanuel Schertzer \\
  Department of Mathematics \\
  University of Vienna, Austria \\
  \texttt{emmanuel.schertzer@univie.ac.at}
}

\newcommand{\francois}[1]{{\color{violet} Francois: #1}}

\newcommand{\EE}{\mathbb{E}}

\newcommand{\NN}{\mathbb{N}}

\newcommand{\RR}{\mathbb{R}}

\usepackage{comment} %%% XXXXXXXX TO REMOVE ! 

\begin{document}

\doparttoc % Tell to minitoc to generate a toc for the parts
\faketableofcontents % Run a fake tableofcontents command for the partocs

\part{} % Start the document part

\maketitle

\begin{abstract}
  Diffusion models now set the benchmark in high-fidelity generative sampling, yet they can, in principle, be prone to memorization. In this case, their learned score overfits the finite dataset so that the reverse-time SDE samples are mostly training points. In this paper, we interpret the empirical score as a noisy version of the true score and show that its covariance matrix is asymptotically a re-weighted data PCA. In large dimension, the small time limit makes the noise variance blow up while simultaneously reducing spatial correlation. To reduce this variance, we introduce a kernel-smoothed empirical score and analyze its bias-variance trade-off. We derive asymptotic bounds on the Kullback-Leibler divergence between the true distribution and the one generated by the modified reverse SDE. Regularization on the score has the same effect as increasing the size of the training dataset, and thus helps prevent memorization. A spectral decomposition of the forward diffusion suggests better variance control under some regularity conditions of the true data distribution. Reverse diffusion with kernel-smoothed empirical score can be reformulated as a gradient descent drifted toward a Log-Exponential Double-Kernel Density Estimator (LED-KDE). This perspective highlights two regularization mechanisms taking place in denoising diffusions: an initial Gaussian kernel first diffuses mass isotropically in the ambient space, while a second kernel applied in score space concentrates and spreads that mass along the data manifold. Hence, even a straightforward regularization—without any learning—already mitigates memorization and enhances generalization. Numerically, we illustrate our results with several experiments on synthetic and MNIST datasets.
\end{abstract}

\section{Introduction}
The goal of diffusion-based generative models is to generate new samples from a target probability distribution \( p_* \), given a finite dataset \( \{x_i\}_{i=1}^N \) of i.i.d.\ samples drawn from it. This is done in two steps: first, the distribution is gradually noised through a diffusion process; then, the process is reversed by following a score function which guides the denoising back toward the original distribution
\cite{sohl-dickstein-2015}.

Memorization refers to a model's tendency to overfit the training data, effectively "memorizing" individual samples rather than learning to generalize from the underlying distribution \cite{vandenburg2021}. 
The problem arises when estimating the score of the reversed diffusion from data. This estimation is commonly formulated as a quadratic minimization problem over the dataset \cite{vincent2011}. 
When this problem is solved exactly, the minimizer is the empirical score function, which by construction, guides the denoising process directly back to the training samples and leads to memorization \cite{Biroli2024}. 

This naturally leads to the central question:
{\it Why do diffusion models generalize well, despite this tendency toward memorization ?} The key lies in the estimation of the \emph{score} function. The solution of the aforementioned quadratic minimization problem is typically approximated by solving the quadratic minimization problem over a parametric model, such as a neural network~\cite{song2021scorebased}. Parametric models inherently introduce a smoothing effect \cite{hastie2009}.
To capture this phenomenon analytically, we adopt a simplifying assumption: the regularizing effect of the parametric model is modeled as a mollification (i.e., a convolution with a smoothing kernel) of the empirical score.

While this approach is admittedly simplistic, we first demonstrate through a toy example that it provides a reasonable depiction of the behavior observed when the empirical score is approximated using a neural network (see Figure \ref{fig:neural-network-smooth}). Moreover, we show that this simplified model offers the advantage of yielding an explicit bias-variance decomposition, thereby revealing how smoothing contributes to promoting generalization.

\subsection{Our contributions}

\textbf{Kernel-smoothed score.} We introduce the mollified score as an estimator of the true score.

\textbf{CLT for empirical score.} We relate the sampling noise to a Gaussian noise in the score (as $N\to\infty$), and study the dimension-dependent covariance explosion rate and decorrelation in the small sampling time limit. 

\textbf{Bias-variance analysis and smaller sampling time.} A bias-variance decomposition of the mollified empirical score shows that it reduces the sampling noise variance without harming the bias. We provide bounds on the KL-divergence between the true distribution and that generated by the diffusion based on the mollified empirical score, showing a faster transition from memorization to generalization than in the diffusion based on the non-regularized empirical score.

\textbf{Spectral viewpoint}. We provide a spectral interpretation of these results in the full-support setting. Taking advantage of the regularity of the data distribution in frequency space suggests that convolution could further reduce variance.

Additional proofs and numerical results, including protocols, can be found in Appendices \ref{ap:proofs} and \ref{ap:numerical}.
 
\section{Related works}

\paragraph{Convergence and generalization.} Significant effort has been dedicated to the study of convergence of diffusion models \cite{chen2023, chenjmlr2023, cui2025preciseasymptoticanalysislearning, boffi2025shallow}. Recently, \cite{strasman2025analysisnoiseschedulescorebased} improved bounds in Wasserstein distance between the target and estimated distributions, and in \cite{li2023on} upper bounds on the KL divergence are derived. \cite{yi2023generalizationdiffusionmodel} studied the generalization of a generative model through a mutual information measure.

\paragraph{Memorization.} In \cite{Biroli2024} (extended in \cite{george2025analysisdiffusionmodelsmanifold}), the score is trained optimally in high-dimension and large data regimes. There is a collapse timescale where the generated samples are attracted to the training points. Memorization has also been documented in pretrained diffusion models, both in unconditional and conditional models \cite{somepalli2023diffusion, carlini2023, somepalli2023understanding}, in particular when the training set size is smaller than the model capacity \cite{yoon2023diffusion,gu2025on}. 
Using statistical physics tools, in a regime of high-dimension, \cite{achilli2024losingdimensionsgeometricmemorization} relate gaps in the spectrum of the score's Jacobian with loss of dimension, corresponding to a memorization phenomenon.
By analyzing the covariance of the noise due to the data set sampling, we get a local PCA that aligns with the data and whose spectrum is related to the score's Jacobian.

\paragraph{Mitigating memorization issues.} The influence of inductive biases of neural networks to learn the score has been studied: \cite{kadkhodaie2024generalization} considers the U-Net \cite{unet}, noting it tends to learn harmonic bases, \cite{li2024understanding} shows empirically a bias towards Gaussian structures and
\cite{kamb2024} seeks simple inductive biases given by locality and equivariance. Another way to mitigate memorization is to modify the model's training, e.g: 
\cite{daras2023ambient} trains on corrupted data, \cite{lyu2022acceleratingdiffusionmodelsearly} stops the forward diffusion process before it reaches a Gaussian distribution and \cite{chen2024memorizationfreediffusionmodels} introduces targeted guidance strategies.

More recently, regularization of the score has been studied:  \cite{taheri2025regularizationmakediffusionmodels} considers a $\ell_1$-regularization of the diffusion loss, in \cite{wibisono2024optimalscoreestimationempirical} an estimator of the score is built based on a Gaussian smoothed measure and in \cite{baptista2025memorizationregularizationgenerativediffusion}, they proved that a closed-form minimizer (in the deterministic flow) leads to memorization and then different regularization techniques are proposed.

Recently, concurrent to our work, \cite{scarvelis2025closedformdiffusionmodels} introduced an smoothed empirical score, showing that the model generalizes on various empirical experiments.
Their work is mostly empirical and does not assess memorization from the generated samples.
We are concerned with the theoretical guarantees of smoothed score estimators on generalization and memorization.
On the theory side, \cite{chen2025interpolationeffectscoresmoothing} studies the generalization ability of score estimator on a one-dimensional mathematically tractable toy model. We study a similar smoothed score estimator in a general setting that includes random high-dimensional data lying on a low-dimensional manifold. We derive bounds on the KL divergence between the measure generated from the smoothed estimator and the true distribution.

%\textbf{Learning diffusion models:} 

%\cite{li2023on} provide 

\section{Mathematical Background}
\label{eq:mb}

\textbf{Forward-Backward Diffusions. \cite{song2021scorebased}}  
Let $p_*$ be a probability distribution on $\RR^d$.
The goal of diffusion-based generative models is to sample from \( p_* \), given a finite dataset \( \{x_i\}_{i=1}^N \) of i.i.d.\ samples drawn from it.
%We define the \emph{empirical measure} associated with these data points as
%\[
%p^N_* = \frac{1}{N} \sum_{i=1}^N \delta_{x_i}.
%\]
The first step of the diffusion process is to add noise to the data by considering the stochastic differential equation
\begin{equation}
\label{eq:diff-forward}
dX_t = \sigma\, dB_t,
\end{equation}
with initial condition $ X_0 \sim p_*$, where \( B_t \) denotes a standard Brownian motion.  
For simplicity, we assume without loss of generality that \( \sigma = 1 \). 
Although more general noising procedures exist—such as the Ornstein-Uhlenbeck process—we restrict our attention to the Brownian motion case for simplicity.
We denote the law of \( X_t \) by \( p_t := \mathcal{L}(X_t) \), so that \( p_0 = p_* \). The first key idea is that for sufficiently large times \( T \), the distribution \( p_T \) becomes close to a centered Gaussian with variance \( T \). In essence, the noising procedure drives the data distribution toward a simple, structureless distribution—effectively erasing information about the original data distribution \( p_* \). This sets the stage for the reverse (denoising) process, which aims to reconstruct samples from \( p_* \). For $T>0$, we define 
the reversed time process $({Y}_{t})_{t\in [0,T]}$ satisfying the SDE
\begin{equation}
d {Y}_{t}={s_{T-t}}({Y}_{t})dt+ d \bar{B}_{t},\qquad {Y}_{0}\sim p_{T} \approx{\cal N}(0,T). \label{eq:backward-equation-true-score}
\end{equation}
where $\bar{B}$ is a standard Brownian motion 
and 
$
{s_{t}}=\nabla\log p_{t}$ is referred to as the {\it true
score}. Considering the Fokker-Plank equations associated to \eqref{eq:diff-forward} and \eqref{eq:backward-equation-true-score}, one shows that $\mathcal{L}(Y_t)=\mathcal{L}(X_{T-t})$ \cite{Anderson1982}. Hence, sampling from $p_*$ can be obtained by sampling from $Y_T$. 

{\bf Learning the score.}
The unknown true score $s_{t}$ is related to $p_*$ via
Tweedie's formula \cite{Robbins1992} 
\begin{equation}
\label{eq:s-m}
s_t(x) \ = \ -\frac{x-m_t(x)}{t}, \ \ \mbox{where
$m_t(x):=\mathbb{E}_{X_0\sim p_* }\left[X_{0}\mid X_{t}=x\right]$}.
\end{equation}

%where the latter term in the integral means an expected value w.r.t $X\sim{\cal G}_t(x,y)dy$. On the one hand, the previous expression allows to express the score through
%Tweedie's formula \ref{REF} 
%\begin{equation}
%\label{eq:s-m}
%s_t(x) \ = \ -\frac{x-m_t(x)}{t}, \ \ \mbox{where
%$m_t(x):=\mathbb{E}_{X_0\sim p_* }\left[X_{0}\mid X_{t}=x\right]$}.
%\end{equation}
In particular, the estimation of $s$ boils down to the estimation of $m$.
Let $\lambda$ be a positive function on $\mathbb{R}_+$. 
Assuming that $m\in L^2(\lambda(t)dt \otimes dx)$, finding $m$ amounts to solving the minimization problem
\begin{equation*}
m \ := \ \mbox{argmin}_{f \in L^2(\lambda(t)dt \otimes dx)} \ \  \int_0^T \mathbb{E}_{X_0\sim p_*}\left(||f_t(X_t) - X_0||^2 \right) \lambda(t)dt.
\end{equation*}
In practice, one considers a parametric model $m^\theta_t(x)$ and uses the empirical loss: 
\begin{equation}
\label{eq:min-problem}
\hat \theta :=\mbox{argmin}_{\theta} \ \  \int_0^T \mathbb{E}_{X_0^N\sim p_*^N}\left(||m^\theta_t(X^N_t) - X^N_0||^2 \right) \lambda(t)  dt,
\end{equation}
where $X_t^N$ satisfies the SDE \eqref{eq:diff-forward} with initial condition $p^N_* = \frac{1}{N} \sum_{i=1}^N \delta_{x_i}$. The measure $p^N_*$ is the empirical distribution associated to the data set.

\begin{figure}[h]
    \centering
\includegraphics[width=0.95\textwidth]{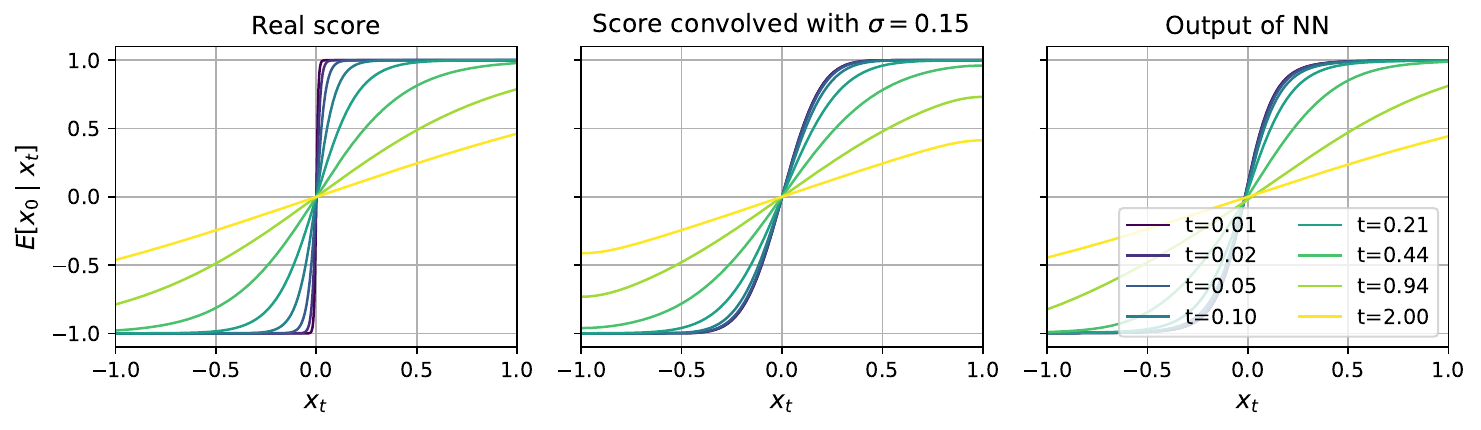}
    \caption{Left: analytical score. Middle: analytical score convolved with a Gaussian kernel with standard deviation $
    \sigma=0.15$. Right: neural network approximation of score. }
    \label{fig:neural-network-smooth}
\end{figure}

{\bf The problem of memorization.} 
Let $p_t^N \ = \ {\cal L}(X^N_t)$ which is the Kernel Density Estimator (KDE) of $p_*$, obtained with a Gaussian kernel with covariance $t\mathrm{Id}_d$.
The \emph{empirical score} $s^N_t := \nabla \log p_t^N$
is defined analogously to the true score, but replacing $p_*$ by its empirical approximation $p_*^N$. By considering $s_t^N$ and $p_T^N$ in place of $s_t$ and $p_T$ in the SDE \eqref{eq:backward-equation-true-score}, the law of the reversed time process now evolves from  $p_T^N\approx {\cal N}(0, \mbox{Id}_d T)$
to the empirical distribution $p_*^N$ at time $T$. As before, 
\begin{equation}
\label{eq:SN}
s^N_t := - \frac{x-m^N_t(x)}{t}, \ \ 
\ \mbox{where
$m^N_t(x):=\mathbb{E}_{X_0\sim p^N_* }\left[X_{0}\mid X_{t}=x\right]$}.
\end{equation}
As $t\to 0$, it is straightforward to see that $m_t^N(x)\to \mbox{argmin}_{i} ||x_i-x||$ so that $m^N$ converges to the
nearest-neighbor map and, since the latter is discontinuous, this shows that the empirical $m^N$ becomes less and less regular. 
%Since the latter is discontinuous, this shows that the empirical $m^N$ becomes less and less regular as $t$ approaches $0$.
See left panel of Fig. \ref{fig:neural-network-smooth}. Intuitively, this pathological behavior relates to the empirical score forcing the diffusion to return to the data set in small time so that generalization can only be achieved 
by an estimated $ m^\theta$ smoothing out those discontinuities.

The next observation is that $m^N$ is solution to the same minimization problem as (\ref{eq:min-problem}) but
replacing the set of candidate functions by $L^2(\lambda(t)dt\otimes dx)$. Memorization happens when the unrestricted minimizer $m^N$ falls (approximately) inside the parametrized family and the optimizer succeeds in finding it, leading to $m^\theta \simeq m^N$. %From the discussion in the previous paragraph,
Thus, memorization can be mitigated in two ways (i) %by the standard observation that 
the choice of parametric space can exclude or penalize non-regular functions, or (ii) the effective numerical resolution of the minimization problem is achieved at a regular solution. 
%\textcolor{red}{XXX : do not agree with that, e.g. when the inductive bias of deep neural network denoisers is aligned with the true distribution \cite{kadkhodaie2024generalization}. }

%{\color{red} define generalisation }

{\bf Mollified score.} While understanding the smoothing effect of the parametric estimation 
(\ref{eq:min-problem}) is presumably a complicated problem, Fig. \ref{fig:neural-network-smooth} suggests that this effect can be captured by a convolution of the empirical score. In this two-point toy model, the analytic score is a $tanh$ with increasing slope as $t\to0$, whereas the learned network has a smoothing effect (right panel)
that is very similar to the convolution of the empirical score (middle panel). Furthermore, wide neural networks in the NTK regime \cite{jacot2018, deepmindpaper} learn a kernel projection of
the empirical score, which can be thought as a kernel convolution (in space and time)
with the NTK's equivalent kernel \cite{rasmussen}. 

In the following, $K(x,y)$ is a kernel and we define the \emph{mollified  score} as
\begin{equation*}
\tilde s^N_t(x):= K\star s^N_t(x) = \int K(x,y)s_t^N(y)dy. 
\end{equation*}
When $\int y K(x,y)  dy= x$, e.g. if $K$ is Gaussian, $\tilde{s}^N_t = -\frac{x-\tilde m^N_t(x)}{t}$, where $\tilde m^N_t(x) = K\star m^N_{t}(x)$.

We will denote by $\tilde Y$ the reversed process associated to the mollified 
empirical score
\begin{equation}
d \tilde {Y}^N_{t}={\tilde s^N_{T-t}}(\tilde {Y}^N_{t})dt + d \bar{B}_{t},\qquad \tilde{Y}_{0}\sim {\cal N}(0,\mbox{Id}_{d} T). \label{eq:backward-equation-mollified-score}
\end{equation}

\paragraph{Low-dimensional data manifold.}

Throughout, we assume that $p_*$ is supported on a smooth differentiable manifold $\mathcal{M}$ with dimension $k\leq d$, where $d$ is the dimension of the ambient space.
We further assume that $p_*$ has a smooth density on $\mathcal{M}$ with uniformly bounded second order derivatives, and by a slight abuse of notation, we identify $p_*$ with its density.
    We will sometimes further (explicitly) assume the following:
    \begin{assumption}
    \label{Assumption linear manifold}
    The manifold $\mathcal{M}$ supporting $p_*$ is a $k$-dimensional linear subspace of $\RR^d$.
    \end{assumption}
    This assumption facilitates the proofs, but we believe that as long as $\mathcal{M}$ has bounded curvature and $p_*$ has a smooth enough density on $\mathcal{M}$, our results still hold up to multiplicative constants depending on $p_*$ and the curvature.

\section{Mollified Empirical Score and Log-Exp. Double-Kernel Density Estimator}

%From now on, we consider a fixed kernel $K(x,y)$. The \emph{mollified estimator} of the true score (or \emph{mollified empirical score}) is $\tilde{s}^N_t:= K \star s^N_t $, obtained by smoothing the empirical score $s_t^N$. The generative diffusion \eqref{eq:reverse-sampling} with model $\tilde{s}^N_t$ defines a family of measures $(\tilde{q}_t^N)_{t\in [0,T]}$. 

In the following, we define the Gaussian kernel 
$$
\forall t>0, \quad
 {\cal G}_t(x,y) := \frac{1}{(2\pi t)^{\frac{d}{2}}}\exp(-||x-y||^2/2t).
$$

Since the empirical score is conservative, the mollified estimator inherits this property. Indeed, 
\begin{eqnarray*}
\tilde{s}^N_t  = K\star \nabla \ln p_t^N = \nabla [K\star \ln p_t^N] =
\nabla\left[ \log\left( \frac{1}{Z_t}\exp\left[ K\star\log\left({\cal G}_t\star p^{{N}}_0\right)\right] \right)\right],
\end{eqnarray*} 
where $Z_t=\int_{\mathbb{R}^d} \exp\left[K\star\log\left({\cal G}_t\star p^{{N}}_0)\right(x)\right] dx $ is a renormalization constant. This motivates the following definition.

\begin{defn}
Let $q$ be a probability distribution. Given two kernels $K$ and $L$, where $L$ is strictly positive, define
\[
({K,L})\star q:=\frac{1}{Z}\exp\left[(K\star\log\left(L\star q)\right)(x)\right],
\]
 where $Z$ is the normalizing constant. 
\end{defn}
The previous computation entails that the mollified score $\tilde s^N_t$
is the score associated to the probability density $(K,\mathcal{G}_t)\star p_0^N$. We refer to the latter quantity as 
the Log-Exp. Double-Kernel Density Estimator (LED-KDE) of $p_*$ at time $t$.

\begin{figure}
\centering
\includegraphics[width=0.9\textwidth]{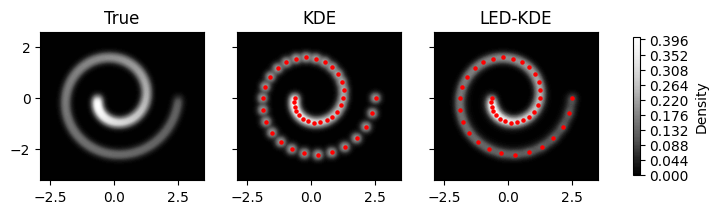}
\caption{Left: True probability measure $p_*$ convolved with a Gaussian kernel with $\sigma=0.02$, $\mathcal{G}_{0.02}$. Middle: KDE with the Gaussian kernel $
\mathcal{G}_{0.02}$. Right: LED-KDE at time $0.02$ with $K=\mathcal{G}_{0.04}$.}
\label{fig:LED-KDE}
\end{figure}

This estimator can be first understood as a two-stage regularization of the empirical measure. The first step is a standard KDE with kernel ${\cal G}_t$, providing initial smoothing and, in a sense, allows to connect data points (this is related to the forward diffusion process). The second step, related to the learned or enforced regularization in the backward diffusion, is a kernel smoothing with kernel $K$, acting in the log-density space to refine the estimator. This mitigates sharp peaks in the KDE estimation since this second regularization does a geometric averaging rather than an arithmetic one. As shown in Fig. \ref{fig:LED-KDE}, we observe that the LED-KDE $(K,{\cal G}_t)\star p_0^N$ provides a much better approximation of the true distribution ${\cal G}_t\star p_*$ as compared to ${\cal G}_t\star p_0^N$.

When the data belongs to a linear subspace, and $K = \mathcal{G}_{\sigma^2}$, the second kernel smoothing in log-density space acts on the KDE by performing smoothing along the data manifold. %\textcolor{red}{since people still put the empirical score in 2025 as a theorem, this could be a proposition}
\begin{proposition}
\label{prop:LED-KDE}
Suppose that Assumption \eqref{Assumption linear manifold} holds and that $\mathcal{M}=\operatorname{span} \{e_1,\dots,e_k \}\subset \mathbb{R}^d$ wlog. Let $\mathcal{G}_{t}^{\mathcal{M}}$ be the Gaussian kernel $\mathcal{N}(0,t\mathrm{Id}_{k}\oplus0_{d-k})$. The measure $(\mathcal{G}_{\sigma^{2}}^{\mathcal{M}},\mathcal{G}_{t}^{\mathcal{M}})\star p_{0}^{N}$ is supported on $\mathbb{R}^k$  and the LED-KDE factors as 
\begin{equation}
(\mathcal{G}_{\sigma^{2}},\mathcal{G}_{t})\star p_{0}^{N}=\left[(\mathcal{G}_{\sigma^{2}}^{\mathcal{M}},\mathcal{G}_{t}^{\mathcal{M}})\star p_{0}^{N}\right]\otimes\mathcal{N}(0,t\mathrm{Id}_{d-k}).
\end{equation}
where on the RHS the first measure is interpreted as a measure on $\mathbb{R}^k$.
\end{proposition}
The RHS of the previous identity can be understood as follows. The data points along the manifold are smoothed out through a LED-KDE on the low dimensional manifold ${\cal M}$, with no leakeage in the ambient space. 
The resulting estimator is then inflated by a Gaussian in the ambient space. This is to be compared with the sole action of ${\cal G}_t$ on $p_0^N$ that directly inflates each data points 
in the ambient space with no prior regularization. Hence, the regularization along the linear manifold induced by the LED-KDE allows one to choose a bandwidth $\mathcal{G}_t$-that would otherwise be considered suboptimal as compared to a KDE. Another consequence is that we can consider a smaller sampling time when using the mollified score, thus reducing the initial mass leakage, without falling into memorization.

%{\color{red} to be rephrazed} {\color{blue} One consequence of this observation, we will show that, compared to the empirical score, with the mollified score, we can consider a smaller sampling time, thus reducing the significant mass leakage, without falling into memorization.}

%[XXX : besoin de mettre le nom en entier LED-KDE ? et Z normalization constant ?]
%As with standard KDE which often uses a bandwidth parameter, the LED-KDE can be implemented with two independent bandwidth
%parameters, one for each kernel. Similarly, the question of selecting the optimal pair of kernels arises, as different choices can impact the quality of the estimator (see Figure XXX).

The fact that the mollified score is the score function of the LED-KDE itself allows us to provide an interpretation of the dynamics \eqref{eq:backward-equation-mollified-score} with the mollified empirical score. Using Otto's formalism \cite{villani, otto1998, bolte2024}, the associated Fokker-Plank equation can be, at least formally, seen as a Wasserstein gradient flow (Appendix \ref{ap:proofs}):
\begin{equation*}
\frac{d}{dt} {\cal L}(\tilde Y^N_{t})=-\frac{1}{2}\mathrm{grad}_{\mathcal{W}} \left(D_{\mathrm{KL}}({\cal L}(\tilde Y^N_{t}) \mid\mid \tilde \mu^N_{T-t}\right),
\end{equation*}
where $\tilde \mu^N_{t} = (2K,\mathcal{G}_t )\star p_0^N$ is a LDE-KDE. Using the empirical score leads to similar equation, with $\mu^N_t = (2\delta_{x=y}, \mathcal{G}_t )\star p_0^N$, essentially a KDE estimation of $p_0^N$. Hence, during the generative dynamics with mollified empirical score, the sampled measure is attracted to a measure which is smoother (along the manifold) than a simple KDE. This provides a first intuition regarding the type of measure that regularized diffusion aim to generate and thus the effect of regularizing the score. 
%In the following, we study the effect of regularization based on a bias-variance study of the empirical and mollified score. This allow us to support the geometric intuition provided in this section. 
%\begin{defn}
%Let $p$ be a probability measure with density on $\mathbb{R}^{d}$. Let $K$ be a kernel. Assuming that either $p$ is positive
%or $K$ is non-negative, the Log-Exponential Kernel Smoothing (LEKS)
%of $p$ is 
%\[
%\frac{1}{Z}\exp\left[(K\star\log p)(x)\right]dx,
%\]
% where $Z = \int \exp\left[(K \star \log p)%(y)\right]dy$ is the
%normalizing constant. 
%\end{defn}
%[Paragraphe : Revenir LED-KDE et aux diffusions] 
%[Paragraphe : Et dire qu'on va parler CLT]
\section{Generative Diffusion and Score Convolution: a bias-variance study. }

We view $m^N$ as a noisy version of the ground truth signal $m$.
Using a CLT on $m^N$, we study the covariance structure of the sampling noise at small times. Using a bias-variance decomposition of the LED-KDE score, we derive asymptotic bounds on the KL divergence of the generated distribution.

\subsection{Sampling Noise, CLT and Re-Weighted PCA}

We write $\overset{\mathrm{f.d.}}{\longrightarrow}$ for a convergence in finite-dimensional distribution.
    Let $G$ be a Gaussian process from $\RR_+\times\RR^d$ to $\RR^d$, with mean zero and covariance matrix at $((t,x),(t',x'))$ given by
    \begin{align*}
        \Sigma_{(t,x),(t',x')}
        &=\EE_{X\sim p_*}\left[(X-m_t(x))(X-m_{t'}(x'))^{\mathrm{T}}\frac{e^{-\frac{\Vert x-X\Vert^2}{2t}}e^{-\frac{\Vert x'-X\Vert^2}{2t'}}}{\EE\big[e^{-\frac{\Vert x-X\Vert^2}{2t}}e^{-\frac{\Vert x'-X\Vert^2}{2t'}}\big]}\right]\times\mathcal{N}_t(x,x'),
    \end{align*}
    where 
    \begin{align*}
        \mathcal{N}_t(x,x')
        :=\frac{\EE_{X\sim p_*}\big[e^{-\frac{\Vert x-X\Vert^2}{2t}}e^{-\frac{\Vert x'-X\Vert^2}{2t'}}\big]}{\EE_{X\sim p_*}\big[e^{-\frac{\Vert x-X\Vert^2}{2t}}\big]\EE_{X\sim p_*}\big[e^{-\frac{\Vert x'-X\Vert^2}{2t'}}\big]}.
    \end{align*}
    The term $\mathcal{N}_t(x,x')$ can be interpreted as the ratio between the expected effective number of points used to estimate the score at both $x$ and $x'$, and the expected effective number of couples $(X,X')$ used to estimate the score at $x$ and $x'$.
        %In appendix, we show another interpretation of the expectation as the conditional expectation of the covariance of $X_0$ given $X_t$.

    For all $x\in\RR^d$ with a unique orthogonal projection onto $\mathcal{M}$, let $\pi(x)$ be that projection, let $T_{\mathcal{M}}(x)\subset\RR^d$ be the tangent space of $\mathcal{M}$ at $\pi(x)$, and let $P_{T_{\mathcal{M}}(x)}:\RR^d\to T_{\mathcal{M}}(x)$ be the orthogonal projection onto $T_{\mathcal{M}}(x)$.
    Under Assumption \ref{Assumption linear manifold}, we write $P_{\mathcal{M}}$ for the orthogonal projection onto $\mathcal{M}$.
    
\begin{thm}\label{thm: covariance asymptotics t to 0}
    (i) The estimator $m^N_t(x)$ is asymptotically normal. More precisely, as $N\to\infty$,
    \begin{align}\label{eq: CLT for m_N}
        \sqrt{N}(m^N_t(x)-m_t(x))&\overset{\mathrm{f.d.}}{\underset{N\to\infty}{\longrightarrow}}
            G(t,x).
        \end{align}
        (ii) Let $t\in(0,\infty)$ and $x\in\RR^d$ with $\pi(x)\in \mbox{Supp}\left(p_*\right)$. Then, it holds that
        \begin{align*}
            \Sigma_{(x,t),(x,t)}
            &\underset{t\to 0}{\sim} \frac{1}{p_*(\pi(x))}\frac{1}{(2\pi)^{k/2}}\frac{1}{t^{k/2-1}}T_{\mathcal{M}}(x).
        \end{align*}
        
        Moreover, under Assumption \ref{Assumption linear manifold}, for all $x_1,x_2\in\RR^d$ such that $\pi_i:=\pi(x_i), \frac{\pi_1+\pi_2}{2}\in \mbox{Supp}(p_*)$ , it holds that
        \begin{align*}
            \Sigma_{(x_1,t),(x_2,t)}
            &\underset{t\to 0}{\sim} e^{-\frac{\Vert\pi_1-\pi_2\Vert^2}{4t}}\frac{p_*\left(\frac{\pi_1+\pi_2}{2}\right)}{p_*(\pi_1)p_*(\pi_2)}\frac{1}{(2\pi)^{k/2}}\frac{1}{t^{k/2-1}}\Big(P_{\mathcal{M}} - \frac{1}{4}(\pi_1-\pi_2)(\pi_1-\pi_2)^{\mathrm{T}}\Big).
        \end{align*}
\end{thm}

The last statement suggests that under Assumption \ref{Assumption linear manifold}, the sampling noise at $x_1$ and $x_2$ with $x_1\neq x_2$ decorrelates as $t\to 0$, with explicit asymptotics on the correlation lengths (see Appendix \ref{ap:proofs}).

Further, the eigenvectors of the covariance matrix $\Sigma_{(t,x),(t,x)}$ yield a local PCA of the data, seen from the point of view of $x$. The projection appearing in the asymptotic behavior of $\Sigma_{(x,t),(x,t)}$ shows that for small $t$, the eigenvectors of the matrix  align with the data.
In particular, the only data noise is in directions tangential to the manifold.
We numerically illustrate this in Figure \ref{fig:SRcovariance} on the Swiss roll dataset (the MNIST dataset can be found in Appendix \ref{ap:numerical}). This behavior is supported in Appendix \ref{ap:proofs} by the fact that, up to small term, $\Sigma_{(x,t),(x,t)}$ is the covariance of $X_0$ conditionally on $X_{\frac{t}{2}}=x$ which itself is related to the score's Jacobian. \cite{kadkhodaie2024generalization} showed how this Jacobian encodes the data manifold. 

\begin{figure}[h]
    \centering
\includegraphics[width=0.80\textwidth]{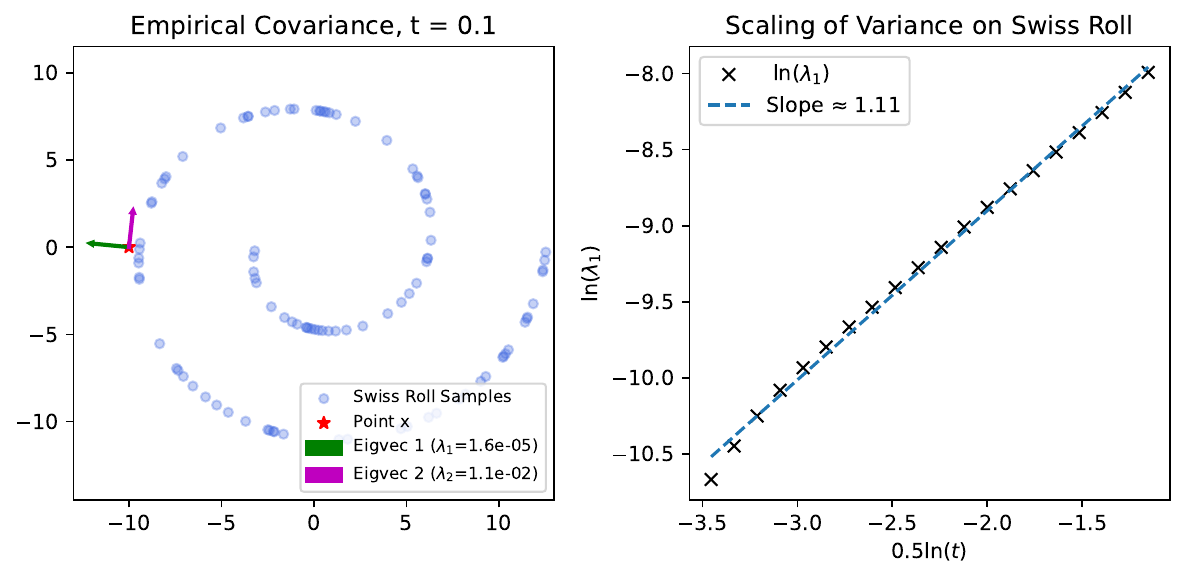}
    \caption{Left: Eigenvector with non-zero corresponding eigenvalue aligned with the data manifold. Right: Scaling of the eigenvalue $\lambda_1$ of empirical covariance matrix ($N=10000$). The slope encodes the intrinsic dimension of the manifold.}
    \label{fig:SRcovariance}
\end{figure}

\subsection{Bias-variance study}
Motivated by our CLT,  we now replace the empirical score $s_t^N$ by its Gaussian approximation
\begin{align}\label{eq: Gaussian approx}
    m_t^{N,G}(x):= m_t(x) + \frac{1}{\sqrt{N}}G(t,x), \ \ \  s_{t}^{N,G}(x) := -  \frac{x-m^{N,G}_{t}(x)}{t}.
\end{align}
For the sake of clarity,
we will abuse notation
and drop the $G$ superscript and use the same definition as Section \ref{eq:mb}.
For instance, we will write $\tilde s_{t}^{N} = \mathcal{G}_h\star s_{t}^{N,G}$, and we stress that the results below are valid up to the validity of the CLT.
For the rest of the paper, we consider the mollified score with the Gaussian kernel $\mathcal{G}_h$.

%$$
%K_h(x,y):=\frac{1}{(2\pi h)^{d/2}}e^{-\frac{\Vert x-y\Vert_2^2}{2h}}.
%$$ \textcolor{red}{Should we put everywhere the $\mathcal{G}_h$} ?}
%The estimator is $\tilde{s}_t^N := K_h \star s_t^N = \frac{1}{t}((K\star m_N(t,\cdot))(x)-x)$. Hence, studying the bias-variance decomposition of $\tilde{s}_t^N$ reduces to analyzing that of $K\star m_N(t,\cdot)$.
%[this will be said before de facon plus voyante/claire]
%Let $t,h\in(0,\infty)$ and let $K_h:x\mapsto \frac{1}{(2\pi h)^{d/2}}e^{-\frac{\Vert x\Vert_2^2}{2h}}$.
%Following definition \francois{LED-KDE}, define the following LED-KDE:
%\begin{align}\label{eq: LED-KDE estimator for diffusion}
%    \widetilde{p}_{t,h}^N
%    &:=\frac{1}{Z_{t,h}}\exp\left(K_h\star\log(K_t\star %p_0^{N})\right)\mathrm{d}x.
%\end{align}
%Let $x\in\RR^d$ be fixed and note that the score of $p_{t,h}^N$ is given by $(K_h\star s_t^N)(x) = \frac{1}{t}((K\star m_N(t,\cdot))(x)-x)$.
We denote $\EE_D[\cdot]$ as the expectation over the dataset $D\!=\!\{x_i\}_{i=1}^N$ composed of i.i.d. random points distributed according to $p_*$. The bias-variance decomposition at $t>0,x\in\RR^d$ yields: %\textcolor{red}{Should we put just a footnote, or to say our bias-variance upper bound is, or... ?}{\color{red} should it be $\tilde m_N$ ?}\francois{I would not} 
\begin{align}\label{eq: bias-variance decomposition}
        \EE_D\Big[\Vert \tilde{m}_t^N(x) - m_t(x)\Vert^2\Big]
        \leq 2\bigg(\underbrace{\EE_{D}\Big[\Vert \tilde{m}_t^N(x) - \tilde{m}_t(x) \Vert^2\Big]}_{v_N(t,h,x)} + \underbrace{\Vert \tilde{m}_t(x) - m_t(x)\Vert^2}_{b(t,h,x)}\bigg)
        %&\hspace{1cm}= \underbrace{\EE_{D}\Big[\Vert (K_h\star m_N(t,\cdot))(x) - (K_h\star m(t,\cdot))(x) \Vert^2\Big]}_{v_N(t,h,x)} + \underbrace{\Vert (K_h\star m(t,\cdot))(x) - m(t,x)\Vert^2}_{b(t,h,x)}\\
        %&\hspace{2cm} - \underbrace{\big\langle (K_h\star m(t,\cdot))(x) - (K_h\star m_N(t,\cdot))(x) , (K_h\star m(t,\cdot))(x) - m(t,x) \big\rangle}_{E(t,h,x)}.
    \end{align}    
\begin{thm}\label{thm: bounds bias-variance from CLT}
     Define $C(x):=\frac{k}{(2\pi)^{\frac{k}{2}}} \frac{1}{p_*(\pi(x))}$. Let $h_N\gg t_N>0$ with $h_N\underset{N\to\infty}{\longrightarrow}0$. Under Assumption \ref{Assumption linear manifold},
     
    (i) 
    If $x\in\RR^d$ is such that $\pi(x)\in \mbox{Supp}(p_*)$, then
            $v_N(t_N,h_N,x)        \underset{N\to\infty}{\longrightarrow} C(x)\frac{t_N}{Nh_N^{\frac{k}{2}}}$.
        %where $C(x)=\frac{k}{(2\pi)^{\frac{k}{2}}\sqrt{k^2-1}}\times\frac{1}{f_*(P_{\mathcal{M}}x)}$.
        
    (ii)
    $b(t_N,h_N,x)\leq d^3 \min\{h_N,h_N^2\} + O(h_Nt_N^2)$.
\end{thm}

One interesting consequence can be obtained by considering the l.h.s. of \eqref{eq: bias-variance decomposition}. It is minimised at the order
%One interesting consequence of Theorem \ref{thm: bounds bias-variance from CLT} can be straightforwardly drawn by optimising $h$ in the upper bound of the bias-variance decomposition: it is minimised at
%Then one can write  
%\begin{align*}
%    h_*:=\argmin_{h>0}\left(d^3h^2 + \frac{C(x)t}{Nh^{\frac{k}{2}}}\right).
%\end{align*}
%and deduce that
\begin{align*}
    h_*=O\left(t/N\right)^{\frac{2}{k+4}},
\end{align*}
showing that as $t\to0$, one needs to reduce the bandwith to reduce the expected $L_2$ error at $(t,x)$. 

\begin{comment}
\begin{wrapfigure}[14]{l}{0.45\textwidth}
  \begin{center}
\vspace{-15pt}
\includegraphics[width=0.45\textwidth]{dropbox_article/images/bias-variance_new_0_morepts.pdf}
  \end{center}
  \vspace{-10pt}
  \caption{Bias-Variance decomposition as in \eqref{eq: bias-variance decomposition} at $x=(-10,0)$ for different $t$ on the Swiss Roll, with $N=300$ and $h=0.1$.}
    \label{fig:bias-varianceSR}
\end{wrapfigure}
\end{comment}

\begin{comment}
In Figure \ref{fig:bias-varianceSR}, we numerically evaluate the bias and the variance on the 2-dimensional Swiss roll test case, considering a point close to the data manifold. {\color{red}comment...}

\end{comment}
\subsection{Enhanced performances and effective data-set size of the mollified score}

We denote by $\tilde q^N_{t,h} \equiv \tilde{q}^N_{t} = {\cal L}(\tilde Y^N_{T-t})$ and $q^N_t = {\cal L}(Y^N_{T-t})$ where the processes $\tilde{Y}^N,{Y}^N$ are the reversed diffusions generated as in \eqref{eq:backward-equation-mollified-score} with $\tilde{s}_t^{G,N}$, respectively $s_t^{G,N}$.

One important tool to capture the problem of generalization was given by
\cite{song2021scorebased} who showed an upper bound on the KL divergence between the smoothed data distribution $p_{t}$ and the generated one $\tilde{q}^N_{t}$: \begin{eqnarray}
D_{KL}\left(p_{t}\left\Vert \tilde {q}^N_{t}\right.\right)\leq\frac{1}{2}\mathbb{L}_t(\tilde s^N) \ + \ D_{KL}\left(p_{T}\left\Vert \mathcal{N}(0, \mathrm{Id}_d T )\right.\right), \nonumber\\
\mbox{where } \ \
\ \mathbb{L}_t(\tilde s^N) := \int_{t}^T \mathbb{E}_{X_u\sim p_u}\left( || s_u(X_u) - \tilde s_u^N(X_u)||^2\right) du. 
\label{eq:KLinequality} 
\end{eqnarray}

Thanks to \eqref{eq:KLinequality} and the bias-variance decomposition of Theorem \ref{thm: bounds bias-variance from CLT}, we obtain asymptotic bounds of the KL divergence at small times $t$ between the true distribution and the generated distributions with and without regularization of the empirical score.

\begin{thm}\label{thm: KL regimes}
    Let $t_N$ and $h_N$ be such that $h_N\gg t_N>0$ with $h_N\underset{N\to\infty}{\longrightarrow}0$. Under Assumption \ref{Assumption linear manifold}, 
\begin{align}
 \EE_D\left[D_{\mathrm{KL}}(p_{t_N}\Vert q_{t_N}^N)\right]
            &\leq O\left(\frac{1}{Nt_N^{\frac{k}{2}}}\right) + D_{\mathrm{KL}}(p_{T}\Vert \mathcal{N}(0,T\mathrm{Id}_d)),\label{thm bound KL empirical}
            \\
            \EE_D\left[D_{\mathrm{KL}}(p_{t_N}\Vert \widetilde{q}_{t_N}^N)\right]
            &\leq O\bigg(\frac{h_N^2}{t_N} + \frac{\log 1/t_N}{Nh_N^{\frac{k}{2}}}\bigg) + D_{\mathrm{KL}}(p_{T}\Vert \mathcal{N}(0,T\mathrm{Id}_d)).\label{thm bound KL regularized}
\end{align}
\end{thm}

    %Memorization happens when sampling from $q_{t_N}^N$ with $t_N$ is too small.
    %Comparing \eqref{thm bound KL empirical} and \eqref{thm bound KL regularized}, we see that the mollified $\widetilde{q}_{t_N,h_N}^N$ can avoid memorization for smaller $t_N$ than $q_{t_N}^N$. {\color{red} say more}
    %this transition from $t_N\gg N^{-\frac{2}{k}}$ to $t_N\gg N^{-\frac{2}{\beta k}}$.
   We begin by adopting the point of view introduced in \cite{kadkhodaie2024generalization}, where a small parameter $t_N > 0$ is fixed, and we investigate how the transition between memorization and generalization depends on the sample size. Inequality \eqref{thm bound KL empirical} suggests that, in the absence of smoothing, this transition occurs at the critical sample size 
\[
N_c = t_N^{-\frac{k}{2}},
\]
which corresponds to a blow-up in the right-hand side of the inequality. To build some intuition for this result, a quick computation shows that when $N \ll N_c$, the quantity $m_N$ becomes degenerate and converges to the nearest-neighbor map: the empirical score forces the reversed diffusion process to return the closest data point—effectively resulting in memorization.

    To analyze the effect of smoothing on the critical sample size, we consider the mollified case with a bandwidth of the form $h_N = t_N^\beta$, where $\beta \in (1/2, 1)$. In this setting, the right-hand side of inequality \eqref{thm bound KL regularized} blows up when $N\ll\tilde N_c$ where
\[
\tilde{N}_c := N_c^{\beta} \ll N_c.
\]
This indicates that a suitable choice of the bandwidth $h$ can significantly reduce the critical sample size at which the transition from memorization to generalization occurs, effectively changing its order of magnitude.

\begin{figure}[h]
    \centering
\includegraphics[width=0.45\textwidth]{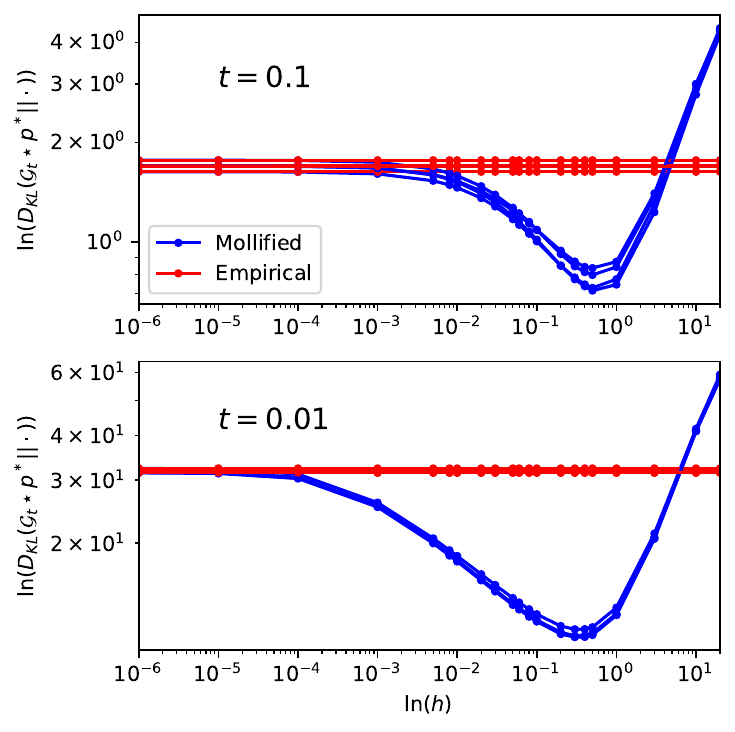}\includegraphics[width=0.45\textwidth]{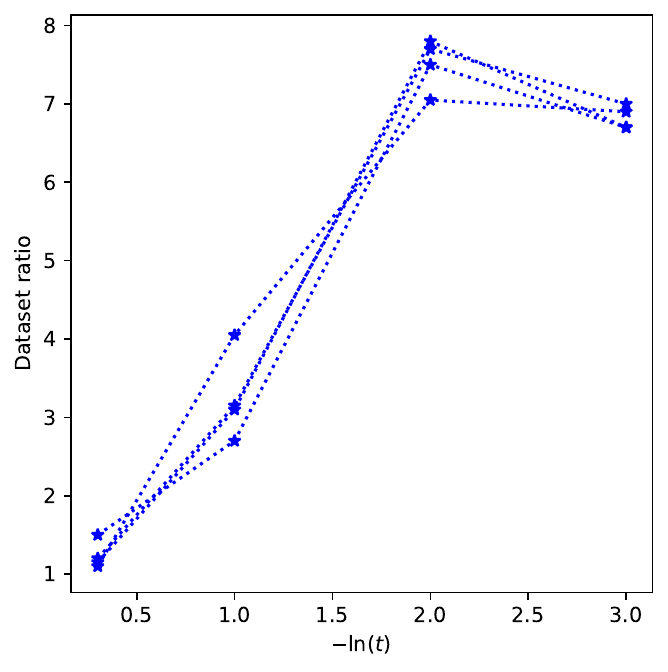}
    \caption{
    %Left: KL-divergence between the empirical measure generated by following the empirical score and the mollified score, varying $h$. 
    Left: KL-divergence between $\mathcal{G}_{t_N}\star p_*$ and the  empirical measure
generated by following the score (red)  and the KL-divergence between  $\mathcal{G}_{t_N}\star p_*$ and the empirical measure generated by following the mollified score, varying $h$  (blue).
    Right: Ratio $N_{\mathrm{eff}}/N$ at the lowest reported KL-divergence. In both figures, $p_*$ is multi-dimensional Gaussian ($d=4$) and  $N=100$.
    }
    \label{fig:effectiveN}
\end{figure}

 Next, define $N_{\mathrm{eff}}$ as
 $$
 \EE_D[D_{\mathrm{KL}}(p_{t}\Vert q_{t}^{N_{\mathrm{eff}}})]
    = \EE_D[D_{\mathrm{KL}}(p_{t}\Vert \widetilde{q}_{t}^N)].$$ 
The upperbounds of the previous theorem suggest that
    $N_{\mathrm{eff}} \approx N (\frac{h}{t})^{\frac{k}{2}}$, which can become very large at small time $t$.
In Figure \ref{fig:effectiveN} (right panel), we numerically estimate $N_{\mathrm{eff}}$ in a toy experiment. The results strongly support the significant improvement of the mollified estimator, especially at very small times, compared to the empirical score with a much larger dataset, showing that for small $t$, $N_{\mathrm{eff}}$ is up to $7\times $ larger than $N$.  On the left, it is shown that a correctly chosen $h$ can lead to a significant decrease of the KL-divergence for the same number of training points. %On the right, we show that for small $t$, up to $7\times$ the amount of training data is necessary to attain the same KL-divergence as when using the mollified score (with the best empirically found $h$). 

\paragraph{Spectral point of view}

We believe that the bounds of Theorem \ref{thm: KL regimes} are not optimal.
In Appendix \ref{ap:proofs}, ignoring the bias term, we develop a heuristic to improve \eqref{thm bound KL regularized} when using an adaptive lengthscale $h=h(t)=t^{\beta}$ at all times, with $\beta\in(0,1)$, that is $\tilde{s}_t^N=\mathcal{G}_{t^{\beta}}\star s_t^N$.
Letting $h_N=h(t_N)=t_N^{\beta}$ for comparison, by leveraging further regularity assumptions on $p_0$ with full support, we obtain
\begin{align*}
    \EE_D\left[D_{\mathrm{KL}}(p_{t_N}\Vert \widetilde{q}_{t_N}^N)\right]
            &\leq O\bigg(\frac{t_N}{Nh_N^{1+\frac{d}{2}}}\bigg) + D_{\mathrm{KL}}(p_{T}\Vert \mathcal{N}(0,T\mathrm{Id}_d)). 
\end{align*}

To obtain this bound, we decompose and study the mollified score in the eigenbasis $f_{\mathbf{k}}=\prod_{j=1}^{d}f_{k_j}(x_j)$, $\mathbf{k}\in\NN^d$ of the Laplacian. Writing $g_{k_j}(x_j)=\partial_{x_j}f_{k_j}(x_j)$, we get 
    \begin{align*}
        (\tilde{s}^N_t)(x)
        &\approx \bigg(\frac{1}{p^N_t(x)}\sum_{\mathbf{k}\in\NN^d}e^{-\pi^2\Vert \mathbf{k}\Vert^2 (t+h)}\frac{-\pi k_m g_{k_m}(x_m)}{f_{k_m}(x_m)}f_{\mathbf{k}}(x)\big\langle p_0^N,f_{\mathbf{k}}\big\rangle\bigg)_{m=1,\ldots,d}.
    \end{align*} 
    Mollification effectively suppresses the high-frequency components of the empirical score (those with $||\mathbf{k}||^2 > O(t + h)^{-1/2}$), which are responsible for its asymptotic degeneracy near the origin.

\section{Discussion}

We study denoising diffusions based on the mollified empirical score, and provide an interpretation based on a two-step smoothing technique -- convolution on the measure, then convolution on the resulting log-likelihood -- to construct a density from an empirical distribution.
Based on the bias-variance decomposition of the mollified empirical score, we show that regularized diffusions are less prone to memorization and have better generalization performances than the non-regularized ones. This translates into a faster transition from memorization to generalization as a function of the dataset size, and enables to preserve good generative performance while decreasing the smallest time of the diffusion, thus reducing the detrimental initial diffusion of mass under the manifold hypothesis.

Even in practice, to avoid memorization, some sort of smoothing must be at play. The present work offers a new perspective to study the generalization of denoising diffusions.
In particular, when the score is approximated by a neural network, say in the neural tangent kernel regime, we conjecture that (part of) the inductive bias could be the result of the kernel convolution of the empirical score with the NTK's equivalent kernel.

Important questions emerge from our analysis: 1. What are the best (possibly time and space dependent) kernels to mollify the score? The covariance matrix $\Sigma$ of Theorem \ref{eq: CLT for m_N} seems to be a good candidate, since it aligns with the data.
2. What is the effect of convolving in space \textbf{and} time?
3. How does the mollified score behave when the higher order terms of the CLT cannot be ignored? 4. How does our analysis compares to other diffusion settings such as with an Ornstein-Uhlenbeck process? We believe the spectral point of view is an interesting lead to extend our approach.

In principle, convolution can be done on any estimator of the score, including neural networks. Since memorization has been reported to occur in practice \cite{somepalli2023diffusion,somepalli2023understanding,carlini2023}, such a regularization technique could be used to mitigate it on a trained network.

\paragraph{Limitations}
1. Our analysis relies on the Gaussian approximation of $m_N$ by the CLT of Theorem \ref{thm: covariance asymptotics t to 0}, which for a fixed $N$ requires the time to not be too small.
2. Even though some linked can be conjectured, we do not consider parametric models of the score such as neural networks used in practice.
3. Our numerical experiments are illustrative of our theoretical results on simple or synthetic settings, but do not attempt at state-of-the-art performance.

\paragraph{Acknowledgments}
E.S. and F.G.G. gratefully acknowledge support from the FWF project PAT3816823. This work was supported in part by an allocation of computing time from the Ohio
Supercomputer Center.

\newpage 
\bibliographystyle{plain}  % Choose a style: plain, ieeetr, apalike, etc.
\bibliography{references}

\newpage
\appendix

\addcontentsline{toc}{section}{Appendix} % Add the appendix text to the document TOC
\part{Appendix} % Start the appendix part
\parttoc

\section{Proofs}
\label{ap:proofs}

\subsection{Assumptions}

As in the main text, $\mathcal{M}:=\mbox{Supp}(\mathcal{M})$ denotes a $k$-dimensional smooth manifold.
We assume $p_*$ has a smooth density on $\mathcal{M}$, such that its second order derivatives are uniformly Lipschitz in $\mathcal{M}$.
For technical reasons, we also assume $p_*(z)>0$ for all $z\in\mathcal{M}$.
We believe this last assumption is superfluous; it guarantees that in the linear manifold case, the orthogonal projection of $x\in\RR^d$ onto $\mathcal{M}$ has positive density under $p_*$, which simplifies the arguments.

\subsection{Proof of Proposition  \ref{prop:LED-KDE}}
\label{sec: proof of proposition LED-KDE}

Recall that $P_{\mathcal{M}}$ and $P_{\mathcal{M}^{\perp}}$ are
the projections on $\mathcal{M}$ and $\mathcal{M}^{\perp}$ respectively.
In the following, we will use the following notation: for any $x\in\mathbb{R}^{d}$,
$x^{(1)}=P_{\mathcal{M}}(x)$ and $x^{(2)}=P_{\mathcal{M}^{\perp}}(x)$.
We denote $\mathcal{G}_{t}^{\mathcal{M}}$ the Gaussian kernel $\mathcal{N}(0,t\mathrm{Id}_{k}\oplus0_{d-k})$
and $\mathcal{G}_{t}^{\mathcal{M}^{\perp}}$ the Gaussian kernel $\mathcal{N}(0,0_{k}\oplus t\mathrm{Id}_{d-k})$. 

Since $\mathcal{G}_{t}(x,x_{i})=\mathcal{G}_{t}^{\mathcal{M}}(x^{(1)},x_{i})\mathcal{G}_{t}^{\mathcal{M}^{\perp}}(x^{(2)},0)$,
we get that $\mathcal{G}_{t}\star p_{0}^{N}=\frac{1}{N}\sum_{i=1}^{N}\mathcal{G}_{t}(\cdot,x_{i})$
is of the form $\left[\mathcal{G}_{t}^{\mathcal{M}}\star p_{0}^{N}\right]\otimes\mathcal{N}(0,t\mathrm{Id}_{d-k}).$
To simplify the notations, let $\mu=\mathcal{G}_{t}^{\mathcal{M}}\star p_{0}^{N}$
and $\nu=\mathcal{N}(0,t\mathrm{Id}_{d-k})$. Then, 
\begin{eqnarray*}
\int_{\mathbb{R}^{d}}\mathcal{G}_{\sigma^{2}}(x,y)\log(\mu\otimes\nu(y))dy & = & \int\mathcal{G}_{\sigma^{2}}^{\mathcal{M}}(x^{(1)},y^{(1)})\mathcal{G}_{\sigma^{2}}^{\mathcal{M^{\perp}}}(x^{(2)},y^{(2)})\log(\mu\otimes\nu(y))dy\\
 & = & \int\mathcal{G}_{\sigma^{2}}^{\mathcal{M}}(x^{(1)},y^{(1)})\mathcal{G}_{\sigma^{2}}^{\mathcal{M^{\perp}}}(x^{(2)},y^{(2)})\log(\mu(y^{(1)}))dy^{(1)}dy^{(2)}\\
 &  & +\int\mathcal{G}_{\sigma^{2}}^{\mathcal{M}}(x^{(1)},y^{(1)})\mathcal{G}_{\sigma^{2}}^{\mathcal{M^{\perp}}}(x^{(2)},y^{(2)})\log(\nu(y^{(2)}))dy^{(1)}dy^{(2)}\\
 & = & \int\mathcal{G}_{\sigma^{2}}^{\mathcal{M}}(x^{(1)},y^{(1)})\log(\mu(y^{(1)}))dy^{(1)}\\
 &  & +\int\mathcal{G}_{\sigma^{2}}^{\mathcal{M^{\perp}}}(x^{(2)},y^{(2)})\log(\nu(y^{(2)}))dy^{(2)}.
\end{eqnarray*}
 Thus, $(\mathcal{G}_{\sigma^{2}},\mathcal{G}_{t})\star p_{0}^{N}(x)$ is proportional to: 
\begin{equation*}
\exp\left(\int\mathcal{G}_{\sigma^{2}}^{\mathcal{M}}(x^{(1)},y^{(1)})\log(\mu(y^{(1)}))dy^{(1)}\right)\exp\left(\int\mathcal{G}_{\sigma^{2}}^{\mathcal{M^{\perp}}}(x^{(2)},y^{(2)})\log(\nu(y^{(2)}))dy^{(2)}\right).
\end{equation*}
 Hence, we obtain: 
\begin{equation*}
(\mathcal{G}_{\sigma^{2}},\mathcal{G}_{t})\star p_{0}^{N}=\left[(\mathcal{G}_{\sigma^{2}}^{\mathcal{M}},\mathcal{G}_{t}^{\mathcal{M}})\star p_{0}^{N}\right]\otimes\tilde{\nu},
\end{equation*}
 where $\tilde{\nu}$ is the probability measure proportional to 
\begin{equation*}
\exp\left(\int\mathcal{G}_{\sigma^{2}}^{\mathcal{M^{\perp}}}(x^{(2)},y^{(2)})\log(\nu(y^{(2)}))dy^{(2)}\right).
\end{equation*}
It remains to show that $\tilde{\nu}=\mathcal{N}(0,t\mathrm{Id}_{d-k})$.
Using the fact that $\nu=\mathcal{N}(0,t\mathrm{Id}_{d-k})$, up to some additive constant which does not depend on $x_0$, 
\begin{equation*}
\int\mathcal{G}_{\sigma^{2}}^{\mathcal{M^{\perp}}}(x^{(2)},y^{(2)})\log(\nu(y^{(2)}))dy^{(2)}  =  \mathbb{E}_{N\sim\mathcal{N}(0,\mathrm{Id}_{d-k})}\left[-\frac{\|x^{(2)}+\sigma N\|^{2}}{2t}\right].
\end{equation*}
Since $\|x^{(2)}+\sigma N\|^{2}=\|x^{(2)}\|^{2}+2\sigma\left\langle x^{(2)},N\right\rangle +\sigma^{2}\|N\|^{2}$,
this is equal to 
$-\frac{\|x^{(2)}\|^{2}}{2t},$
up to an additive constant which does not depend on $x^{(2)}$.
Hence, $\tilde{\nu}$ is the probability measure proportional to $\exp(-\frac{\|x^{(2)}\|^{2}}{2t})$:
it is $\mathcal{N}(0,t\mathrm{Id}_{d-k})$. This allows us to conclude.

\begin{rem}
    This proposition holds true because smoothing in log-density space respects the tensor product. Besides, the use of Gaussian kernels respects Gaussian distributions. Smoothing in log-density space has another interesting property: it shrinks the support of measures instead of putting mass outside the support of the measure to smooth. 
\end{rem}

\subsection{LED-KDE and gradient descent}
\label{sec: LDE-KDE Wasserstein flow}
We first recall the relation between diffusions and Wasserstein gradient
descent, as explained in Section 6.2 of \cite{bolte2024}. Consider the stochastic
differential equation $dX_{t}=-\beta_{t}\nabla U_{t}(X_{t})dt+\sqrt{2}dW_{t}$
and its law $\rho_{t}=\mathcal{L}(X_{t}).$ The family of laws $(\rho_{t})_{t\geq0}$
satisfies the Fokker-Plank equation 
\begin{equation*}
\partial_{t}\rho_{t}=\beta_{t}\mathrm{div}\left(\rho_{t}\nabla U_{t}\right)+\Delta\rho_{t}=\nabla \cdot \left[(\beta_{t}\nabla U_{t}+\frac{\nabla\rho_{t}}{\rho_{t}})\rho_{t}\right].
\end{equation*}
This can then be written as 
\begin{equation*}
\frac{d}{dt}\rho_{t}=-\mathrm{grad}_{\mathcal{W}}\mathcal{U}_{\beta_{t}}(\rho_{t})
\end{equation*}
 where $\mathcal{U}_{\beta}(\rho)=\beta\int U(x)\rho(x)dx+\int\rho(x)\log\rho(x)dx$.
This is the KL-divergence between $\rho$ and the measure with score
$-\beta\nabla U$. 

In our setting, the drift is of the form $\hat{s}_{t}=\nabla\ln\hat{p}_{T-t}$
where $\hat{p}_{T-t}$ is either the KDE measure or the LED-KDE measure
of $p_{*}$ given the dataset, and we consider the SDE: 
\begin{equation*}
dY_{t}=\hat{s}_{t}(Y_{t})dt+dW_{t}.
\end{equation*}
 The corresponding Fokker-Plank equation satisfied by $\rho_{t}=\mathcal{L}(Y_{t})$
is thus 
\begin{equation*}
\partial\rho_{t}=-\nabla \cdot (\hat{s}_{t}\rho)+\frac{1}{2}\Delta\rho=\frac{1}{2}\nabla \cdot \left[\left(-2\hat{s}_{t}+\frac{\nabla\rho}{\rho}\right)\rho\right].
\end{equation*}
Hence we are in the same setup as before, as long as we add the
$\frac{1}{2}$ factor in front and replace $\beta_{t}\nabla U_{t}$
by $-2\hat{s}_{t}$. This yields 
\begin{equation*}
\frac{d}{dt}\rho_{t}=-\frac{1}{2}\mathrm{grad}_{\mathcal{W}}\mathcal{F}(\rho_{t})
\end{equation*}
with $\mathcal{F}_{t}(\rho)=D_{\mathrm{KL}}(\rho\mid\mid\mu_{t}),$
where $\mu_{t}$ is the probability measure with score $2\hat{s}_{t}$. When
$\hat{s}_{t}$ is the score of the KDE measure, this measure is $(2\delta_{x=y},\mathcal{G}_{t})\star p_{0}^{N}$, where $\delta_{x,y}$ is the Dirac kernel. When $\hat{s}_{t}$ is the mollified version with kernel $K$, we have $\mu_{t}=(2K,\mathcal{G}_{t})\star p_{0}^{N}$.

\subsection{Proof of Theorem \ref{thm: covariance asymptotics t to 0}}\label{sec: proof of thm covariance asymptotics}

\begin{proof}[(Proof of Theorem \ref{thm: covariance asymptotics t to 0})]

    (i) We first prove that the estimator $m^N_t(x)$ is asymptotically normal. More precisely, as $N\to\infty$,
    \begin{align}\label{eq: CLT for m_N}
        \sqrt{N}(m^N_t(x)-m_t(x))&\overset{\mathrm{f.d.}}{\underset{N\to\infty}{\longrightarrow}}
            G(t,x).
        \end{align}

\textit{Proof}: Recall that
\begin{equation}
m^{N}_t(x)=\frac{\frac{1}{N}\sum_{i=1}^{N}x_{i}e^{-\frac{\left\Vert x-x_{i}\right\Vert ^{2}}{2t}}}{ \frac{1}{N}\sum_{i=1}^{N} e^{-\frac{\left\Vert x-x_{i}\right\Vert ^{2}}{2t}}}.\label{eq:m_N_equation}
\end{equation}
The general idea is to apply a central limit theorem on the numerator and denominator, followed by a Taylor expansion. In the following, we provide a rigorous way to do so. Fix $t,t'>0$ and $x,x'\in\mathbb{R}^{d}$. Consider the random variable
\begin{equation*}
W=\left(e^{-\frac{\left\Vert x-Z\right\Vert ^{2}}{2t}},Ze^{-\frac{\left\Vert x-Z\right\Vert ^{2}}{2t}},e^{-\frac{\left\Vert x'-Z\right\Vert ^{2}}{2t'}},Ze^{-\frac{\left\Vert x'-Z\right\Vert ^{2}}{2t'}}\right)\in\mathbb{R}^{2(d+1)},
\end{equation*}
 where $Z\sim p_{*}.$ Then 
\begin{equation*}
w_{i}:=\left(e^{-\frac{\left\Vert x-x_{i}\right\Vert ^{2}}{2t}},x_{i}e^{-\frac{\left\Vert x-x_{i}\right\Vert ^{2}}{2t}},e^{-\frac{\left\Vert x'-x_{i}\right\Vert ^{2}}{2t'}},x_{i}e^{-\frac{\left\Vert x-x_{i}\right\Vert ^{2}}{2t}}\right)
\end{equation*}
are i.i.d. samples with same law as $W$. 

For $\epsilon\in\{0,1\}$, and $t,x$ we define
\begin{equation*}
\varphi^{(\epsilon)}(t,x)=\mathbb{E}_{Z\sim p_{*}}\left[Z^{\epsilon}e^{-\frac{\left\Vert x-Z\right\Vert ^{2}}{2t}}\right]. 
\end{equation*}

Using the Central Limit Theorem, along with the Skorokhod representation
theorem, there exist $S_{1},\ldots,S_{N},...$ where
$S_{N}$ has the same law as $\frac{1}{N}\sum_{i=1}^{N}W_{i}$ such
that the convergence 
\begin{equation}
\sqrt{N}\left[S_{N}-(\varphi^{(0)}(t,x),\varphi^{(1)}(t,x),\varphi^{(0)}(t',x'),\varphi^{(1)}(t',x'))\right]\underset{N\to\infty}{\longrightarrow}\mathcal{N}\label{eq:CLT}
\end{equation}
 holds almost surely, and
\begin{equation*}
\mathcal{N}=\left(\psi^{(0)}(t,x),\psi^{(1)}(t,x),\psi^{(0)}(t',x'),\psi^{(1)}(t',x')\right)
\end{equation*}
where $(\psi^{(\epsilon)}(t,x))_{t,x,\epsilon}$ is a centered
Gaussian process whose covariance is given by
\begin{equation*}
\mathbb{E}\left[\psi^{(\epsilon)}(t,x)\psi^{(\epsilon')}(t',x')\right]=\mathrm{Cov}(Z^{\epsilon}e^{-\frac{\mid\mid x-Z\mid\mid^{2}}{2t}},Z^{\epsilon'}e^{-\frac{\mid\mid x'-Z\mid\mid^{2}}{2t'}}).
\end{equation*}

Denote $S_{N}=\left(\Phi_{N}^{(0)}(t,x),\Phi_{N}^{(1)}(t,x),\Phi_{N}^{(0)}(t',x'),\Phi_{N}^{(1)}(t',x')\right)$
where for any $\epsilon\in\{0,1\},$ $t>0,$  and $x\in\mathbb{R}^{d}$, we have the equality in law 
\begin{equation*}
\Phi_{N}^{(\epsilon)}(t,x)\overset{d}{=}\frac{1}{N}\sum_{i=1}^{N}x_{i}^{\epsilon}e^{-\frac{\mid\mid x-x_{i}\mid\mid^{2}}{2t}}.
\end{equation*}
From Equation (\ref{eq:m_N_equation}), $\left(\sqrt{N}\left[m_t^{N}(x)-m_t(x)\right],\sqrt{N}\left[m^{N}_{t'}(x')-m_{t'}(x')\right]\right)$
has the same law as 
\begin{equation*}
\left(\sqrt{N}\left[\frac{\Phi_{N}^{(1)}(t,x)}{\Phi_{N}^{(0)}(t,x)}-m_t(x)\right],\sqrt{N}\left[\frac{\Phi_{N}^{(1)}(t',x')}{\Phi_{N}^{(0)}(t',x')}-m_{t'}(x')\right]\right).
\end{equation*}
From Equation (\ref{eq:CLT}), we have almost surely
\begin{eqnarray*}
\Phi^{(0)}(t,x) & = & \varphi^{(0)}(t,x)+\frac{1}{\sqrt{N}}\psi^{(0)}(t,x)+o(N^{-\frac{1}{2}}),\\
\Phi^{(1)}(t,x) & = & \varphi^{(1)}(t,x)+\frac{1}{\sqrt{N}}\psi^{(1)}(t,x)+o(N^{-\frac{1}{2}}).
\end{eqnarray*}
Since 
\begin{equation*}
m_t(x)=\mathbb{E}\left[Z\frac{e^{-\frac{\mid\mid x-Z\mid\mid^{2}}{2\sigma^{2}t}}}{\mathbb{E}[e^{-\frac{\mid\mid x-Z\mid\mid^{2}}{2\sigma^{2}t}}]}\right] = \frac{\varphi^{(1)}(t,x)}{\varphi^{(0)}(t,x)},
\end{equation*}
we obtain up to order $N^{-\frac{1}{2}},$ 
\begin{eqnarray*}
\frac{\Phi_{N}^{(1)}(t,x)}{\Phi_{N}^{(0)}(t,x)} & = & \frac{\varphi^{(1)}(t,x)+\frac{1}{\sqrt{N}}\psi^{(1)}(t,x)}{\varphi^{(0)}(t,x)+\frac{1}{\sqrt{N}}\psi^{(0)}(t,x)}\\%=\frac{1}{\varphi^{(0)}(t,x)}\left[\varphi^{(1)}(t,x)+\frac{1}{\sqrt{N}}\psi^{(1)}(t,x)\right]\left[1-\frac{1}{\sqrt{N}}\frac{\psi^{(0)}(t,x)}{\varphi^{(0)}(t,x)}\right]\\
 & = & \frac{1}{\varphi^{(0)}(t,x)}\left[\varphi^{(1)}(t,x)+\frac{1}{\sqrt{N}}\psi^{(1)}(t,x)-\frac{\varphi^{(1)}(t,x)}{\sqrt{N}}\frac{\psi^{(0)}(t,x)}{\varphi^{(0)}(t,x)}\right]\\
 & = & \frac{\varphi^{(1)}(t,x)}{\varphi^{(0)}(t,x)}+\frac{1}{\sqrt{N}}\frac{\psi^{(1)}(t,x)}{\varphi^{(0)}(t,x)}-\frac{1}{\sqrt{N}}\frac{\varphi^{(1)}(t,x)}{\varphi^{(0)}(t,x)}\frac{\psi^{(0)}(t,x)}{\varphi^{(0)}(t,x)}\\
 & = & m_t(x)+\frac{1}{\sqrt{N}}\frac{\psi^{(1)}(t,x)-m_t(x)\psi^{(0)}(t,x)}{\varphi^{(0)}(t,x)}.
\end{eqnarray*}
 Hence,
\[
\sqrt{N}\left[\frac{\Phi_{N}^{(1)}(t,x)}{\Phi_{N}^{(0)}(t,x)}-m_t(x)\right]\underset{N\to\infty}{\longrightarrow}\frac{\psi^{(1)}(t,x)-m_t(x)\psi^{(0)}(t,x)}{\varphi^{(0)}(t,x)},
\]
and similarly for $t'$ and $x'$. In particular,
\[
\left(\sqrt{N}\left[m^{N}_t(x)-m_t(x)\right],\sqrt{N}\left[m^{N}_{t'}(x')-m_{t'}(x')\right]\right)
\]
 converges in law to
\[
\left(\frac{\psi^{(1)}(t,x)-m_t(x)\psi^{(0)}(t,x)}{\varphi^{(0)}(t,x)},\frac{\psi^{(1)}(t',x')-m_{t'}(x')\psi^{(0)}(t',x')}{\varphi^{(0)}(t',x')}\right).
\]
Given that $(\psi^{(\epsilon)}(t,x))_{t,x,\epsilon}$ is a Gaussian
process, the process
\[
\eta(t,x)=\frac{\psi^{(1)}(t,x)-m_t(x)\psi^{(0)}(t,x)}{\varphi^{(0)}(t,x)}
\]
is Gaussian. To compute its covariance, we can simply replace
$\psi^{(\epsilon)}(t,x)$ by $Z^{\epsilon}e^{-\frac{\mid\mid x-Z\mid\mid^{2}}{2t}}$,
and thus $\mathrm{Cov}\left[\eta(t,x),\eta(t',x')\right]$
 is equal to 
\begin{equation*}
\mathrm{Cov}\left[(Z-m_t(x))\frac{e^{-\frac{\mid\mid x-Z\mid\mid^{2}}{2t}}}{\mathbb{E}[e^{-\frac{\mid\mid x-Z\mid\mid^{2}}{2t}}]},(Z-m_{t'}(x'))\frac{e^{-\frac{\mid\mid x'-Z\mid\mid^{2}}{2t'}}}{\mathbb{E}[e^{-\frac{\mid\mid x'-Z\mid\mid^{2}}{2t'}}]}\right].
\end{equation*}
By definition of $m_t$ and $m_{t'}$, the two terms are centered thus, this covariance is 
\begin{equation*}
\mathbb{E}\left[(Z-m(t,x))(Z-m(t',x'))^{T}\frac{e^{-\frac{\mid\mid x-Z\mid\mid^{2}}{2t}}}{\mathbb{E}[e^{-\frac{\mid\mid x-Z\mid\mid^{2}}{2t}}]}\frac{e^{-\frac{\mid\mid x'-Z\mid\mid^{2}}{2t'}}}{\mathbb{E}[e^{-\frac{\mid\mid x'-Z\mid\mid^{2}}{2t'}}]}\right]. 
\end{equation*}
This allows us to conclude.

    (ii)
    Recall that
    \begin{align}\label{eq: in proof covarance matrix}
        \Sigma_{(t,x),(t,x')}
        &=\EE_{X\sim p_*}\left[(X-m_t(x))(X-m_{t}(x'))^{\mathrm{T}}\frac{e^{-\frac{\Vert x-X\Vert^2}{2t}}e^{-\frac{\Vert x'-X\Vert^2}{2t}}}{\EE_{X\sim p_*}\big[e^{-\frac{\Vert x-X\Vert^2}{2t}}\big]\EE_{X\sim p_*}\big[e^{-\frac{\Vert x'-X\Vert^2}{2t}}\big]}\right].
    \end{align}
    We first focus on the asymptotic behavior of the denominator as $t\to 0$.
    Since $\mathcal{M}$ is smooth, we have by Laplace's Method (see \cite{bender2013advanced} Chapter 6) that
    \begin{align*}
        \EE_{X\sim p_*}\big[ e^{-\frac{\Vert x-X\Vert^2}{2t}}\big]
        &=\int_{\mathcal{M}}e^{-\frac{\Vert x-z\Vert^2}{2t}}p_*(z)\mathrm{d}z\\
        &=e^{-\frac{\Vert x - \pi(x)\Vert^2}{2t}}\int_{\mathcal{M}}e^{-\frac{\Vert \pi(x)-z\Vert^2}{2t}}p_*(z)\mathrm{d}z\\
        &\underset{t\to 0}{\sim}e^{-\frac{\Vert x - \pi(x)\Vert^2}{2t}}(2\pi t)^{\frac{k}{2}}p_*(\pi(x)).
    \end{align*}
    We deduce that
    \begin{align*}
        \EE_{X\sim p_*}\big[e^{-\frac{\Vert x-X\Vert^2}{2t}}\big]\EE_{X\sim p_*}\big[e^{-\frac{\Vert x'-X\Vert^2}{2t'}}\big]
        \underset{t\to 0}{\sim}e^{-\frac{\Vert x - \pi(x)\Vert^2 + \Vert x' - \pi(x')\Vert^2}{2t}}(2\pi t)^{k}p_*(\pi(x))p_*(\pi(x')).
    \end{align*}
    We now turn to the numerator of \eqref{eq: in proof covarance matrix}.
    We have
    \begin{align*}
        &\EE_{X\sim p_*}\left[(X-m_t(x))(X-m_{t}(x'))^{\mathrm{T}}e^{-\frac{\Vert x-X\Vert^2}{2t}}e^{-\frac{\Vert x'-X\Vert^2}{2t}}\right]\\
        &\hspace{1cm}=e^{-\frac{\Vert x-\pi(x)\Vert^2 + \Vert x'-\pi(x')\Vert^2}{2t}}\int_{\mathcal{M}}(z-m_t(x))(z-m_t(x'))^{\mathrm{T}}e^{-\frac{\Vert \pi(x)-z\Vert^2}{2t}}e^{-\frac{\Vert \pi(x')-z\Vert^2}{2t}}p_*(z)\mathrm{d}z.
    \end{align*}
    One can check that
    \begin{align*}
        \Vert \pi(x)-z\Vert^2 + \Vert \pi(x')-z\Vert^2
        &= 2\Big\Vert z-\frac{\pi(x)+\pi(x')}{2}\Big\Vert^2 + 2\Big\Vert\frac{\pi(x)-\pi(x')}{2}\Big\Vert^2,
    \end{align*}
    which can be plugged in the right-hand side above to write
    \begin{align*}
        &\EE_{X\sim p_*}\left[(X-m_t(x))(X-m_{t}(x'))^{\mathrm{T}}e^{-\frac{\Vert x-X\Vert^2}{2t}}e^{-\frac{\Vert x'-X\Vert^2}{2t}}\right]\\
        &\hspace{0.3cm}=e^{-\frac{\Vert x-\pi(x)\Vert^2 + \Vert x'-\pi(x')\Vert^2 - 2\Vert\frac{\pi(x)-\pi(x')}{2}\Vert^2}{2t}}\int_{\mathcal{M}}(z-m_t(x))(z-m_t(x'))^{\mathrm{T}}e^{-\frac{\Vert \frac{\pi(x)+\pi(x')}{2}-z\Vert^2}{t}}p_*(z)\mathrm{d}z.
    \end{align*}
    If $\mathcal{M}$ is linear, we use a change of variable to write the integral as
    \begin{align*}
        &\int_{\mathcal{M}}\Big(z + \frac{\pi(x)+\pi(x')}{2} - m_t(x)\Big)\Big(z + \frac{\pi(x)+\pi(x')}{2} -m_t(x')\Big)^{\mathrm{T}}e^{-\frac{\Vert z\Vert^2}{t}}p_*\Big(z + \frac{\pi(x)+\pi(x')}{2}\Big)\mathrm{d}z.
    \end{align*}
    Since $m_t(x)\to\pi(x)$ as $t\to 0+$ and similarly for $m_t(x')$, another use of Laplace's Method shows that
    \begin{align*}
        &\EE_{X\sim p_*}\left[(X-m_t(x))(X-m_{t}(x'))^{\mathrm{T}}e^{-\frac{\Vert x-X\Vert^2}{2t}}e^{-\frac{\Vert x'-X\Vert^2}{2t}}\right]\\
        &\hspace{2cm}\underset{t\to 0}{\sim}e^{-\frac{\Vert x-\pi(x)\Vert^2 + \Vert x'-\pi(x')\Vert^2 - 2\Vert\frac{\pi(x)-\pi(x')}{2}\Vert^2}{2t}}(2\pi t)^{k/2}p_*\Big(\frac{\pi(x)+\pi(x')}{2}\Big)\\
        &\hspace{6cm}\times \Big(P_{\mathcal{M}}-\frac{1}{4}(\pi(x)-\pi(x'))(\pi(x)-\pi(x'))^{\mathrm{T}}\Big).
    \end{align*}
    The case where $x=x'$ does not require Assumption \ref{Assumption linear manifold} and follows from the same argument.
    This ends the proof.
\end{proof}

Note that in the following the term $-\frac{1}{4}(\pi(x)-\pi(x'))(\pi(x)-\pi(x'))^{\mathrm{T}}$ will not play an important role, because we shall only use $\Sigma_{(t,x),(t,x')}$ with $x\neq x'$ after the convolution with lengthscale $h\to 0$, so that this term can be neglected.

More importantly, each application of Laplace's Method in the above proof was on integrals of the form $\int e^{-\frac{\Vert z-\pi(x)\Vert^2}{2t}}f(z)p_*(z)\mathrm{d}z$ with $f(z)=1,z$, or $z^2$.
Since $p_*$ is smooth with uniformly Lipschitz second order derivatives on $\mathcal{M}$ (and since $\mathcal{M}$ is a smooth manifold), one has
\begin{align*}
    \int_{\mathcal{M}}e^{-\frac{\Vert z-\pi(x)\Vert^2}{2t}}f(z)p_*(z)\mathrm{d}z
    &\underset{t\to 0}{\sim} (2\pi t)^{\frac{k}{2}}f(\pi(x))p_*(\pi(x)) + O(t^{1+\frac{k}{2}}),
\end{align*}
where the $O$ term is uniform in $x\in\RR^d$.
To see why, consider the case of a linear manifold $\mathcal{M}$ (the asymptotic behavior is the same if $\mathcal{M}$ is not linear as long as it is smooth), for simplicity choose $f(z)=1$, and use a change of variable then Taylor's Theorem to write
\begin{align*}
    &\int_{\mathcal{M}}e^{-\frac{\Vert z-\pi(x)\Vert^2}{2t}}(p_*(z)-p_*(\pi(x))\mathrm{d}z\\
    &\hspace{2cm}= (\sqrt{t})^{k}\int_{\mathcal{M}}e^{-\frac{\Vert z\Vert^2}{2}}(p_*(\sqrt{t}z+\pi(x))-p_*(\pi(x)))\mathrm{d}z\\
    &\hspace{2cm}= (\sqrt{t})^{k}\int_{\mathcal{M}}e^{-\frac{\Vert z\Vert^2}{2}}\sqrt{t}z^{T}\nabla p_*(\pi(x)) + tz^{T}\mathcal{H}_{p_*}(\pi(x) + \lambda_{z}\sqrt{t}z)z\mathrm{d}z,
\end{align*}
for some $\lambda_z\in(0,1)$, where $\mathcal{H}_{p_*}(y)$ denotes the Hessian matrix of $p_*$ at $y$. Since $\Vert \mathcal{H}_{p_*}(\pi(x) + \lambda_{z}\sqrt{t}z)\Vert$ is uniformly bounded by assumption, the right-hand side above is $O(t^{1+\frac{k}{2}})$, which yields the claim.
We refer to \cite{bender2013advanced} for more details on the higher order terms in the Laplace's Method (in particular Equation (6.4.45)).

\subsection{Covariance and Score}
\label{ap:covariance_and_score}
Recall that the covariance matrix is given by 
\begin{equation*}
\Sigma_{(x,t),(x',t')}=\underline{\Sigma}_{(x,t),(x',t')}\times\mathcal{N}_{t}(x,x')
\end{equation*}
where 
\begin{equation*}
\underline{\Sigma}_{(x,t),(x',t')}:=\mathbb{E}_{X\sim p_{*}}\left[(X-m_{t}(x))(X-m_{t'}(x'))^{T}\frac{e^{-\frac{\|x-X\|^{2}}{2t}}e^{-\frac{\|x'-X\|^{2}}{2t'}}}{\mathbb{E}\left[e^{-\frac{\|x-X\|^{2}}{2t}}e^{-\frac{\|x'-X\|^{2}}{2t'}}\right]}\right].
\end{equation*}
We now provide alternative formulations of $\underline{\Sigma}_{(x,t),(x',t')}$. 

\textbf{Formulation as a conditional expectation}: Let $X_{0}\sim p_{*}$,
and $B^{(1)},B^{(2)}\sim\mathcal{N}(0,\mathrm{Id}_{d})$ be
three independent random variables. Define 
\begin{equation*}
X_{t}^{(1)}=X_{0}+\sqrt{t}B^{(1)},\qquad X_{t}^{(2)}=X_{0}+\sqrt{t'}B^{(2)}.
\end{equation*}
Then,
\begin{equation*}
\underline{\Sigma}_{(x,t),(x',t')}=\mathbb{E}\left[\left(X_{0}-\mathbb{E}[X_{0}\mid X_{t}^{(1)}=x]\right)\left(X_{0}-\mathbb{E}[X_{0}\mid X_{t'}^{(2)}=x']\right)^{T}\mid X_{t}^{(1)}=x,X_{t'}^{(2)}=x'\right].
\end{equation*}
Indeed, $m_{t}(x)=\mathbb{E}[X_{0}\mid X_{t}^{(1)}=x]$, and similarly
for $m_{t'}(x')$. Besides 
\begin{eqnarray*}
p(x_{0}\mid x_{t}^{(1)}=x,x_{t'}^{(2)}=x') & \propto & p(x_{t}^{(1)}=x,x_{t'}^{(2)}=x'\mid x_{0})p_{*}(x_{0})\\
 & \propto & e^{-\frac{\mid\mid x-x_{0}\mid\mid^{2}}{2t}}e^{-\frac{\mid\mid x'-x_{0}\mid\mid^{2}}{2t'}}p_{*}(x_{0}).
\end{eqnarray*}
When $(t',x')=(t,x)$, we get also the alternative formulation 
\begin{equation}
\underline{\Sigma}_{(x,t),(x,t)}=\mathbb{E}\left[\left(X_{0}-\mathbb{E}[X_{0}\mid X_{t}=x]\right)\left(X_{0}-\mathbb{E}[X_{0}\mid X_{t}=x]\right)^{T}\mid X_{\frac{t}{2}}=x\right].\label{eq:cov-formula-1}
\end{equation}
\textbf{ Relation with the Jacobian of the score}: The score is given
by 
\begin{equation*}
s_{t}(x)=\mathbb{E}_{X\sim p^{*}}\left[-\frac{x-X}{t} \omega_{t,x}(X)\right].
\end{equation*}
where $\omega_{t,x}(X)=\frac{e^{-\frac{\mid\mid x-X\mid\mid^{2}}{2t}}}{\mathbb{E}\left[e^{-\frac{\mid\mid x-X\mid\mid^{2}}{2t}}\right]}$.
Hence, we obtain the classical formula for the Jacobian of the score: 
\begin{eqnarray*}
\nabla s_{t}(x) & = & \mathbb{E}\left[-\frac{\mathrm{Id}_{d}}{t}\omega_{t,x}(X)\right]+\mathbb{E}\left[\left(\frac{x-X}{t}\right)\left(\frac{x-X}{t}\right)^{T}\omega_{t,x}(X)\right]\\
&  & \qquad - \mathbb{E}\left[\frac{x-X}{t}\omega_{t,x}(X)\right]\mathbb{E}\left[\frac{x-X}{t}\omega_{t,x}(X)\right]^{T}\\
 & = & -\frac{\mathrm{Id}_{d}}{t}+\frac{1}{t^{2}}\left\{ \mathbb{E}\left[\left(X-x\right)\left(X-x\right)^{T}\omega_{t,x}(X)\right]-\mathbb{E}\left[\left(X-x\right)\omega_{t,x}(X)\right]\mathbb{E}\left[\left(X-x\right)\omega_{t,x}(X)\right]^{T}\right\} \\
 & = & -\frac{\mathrm{Id}_{d}}{t}+\frac{1}{t^{2}}\left(\mathbb{E}_{X_{0}\mid X_{t}=x}\left[\left(X_{0}-x\right)\left(X_{0}-x\right)^{T}\right]-\mathbb{E}_{X_{0}\mid X_{t}=x}\left[\left(X_{0}-x\right)\right]\mathbb{E}_{X_{0}\mid X_{t}=x}\left[\left(X_{0}-x\right)\right]^{T}\right)\\
 & = & -\frac{\mathrm{Id}_{d}}{t}+\frac{1}{t^{2}}\mathrm{Cov}_{X_{0}\mid X_{t}=x}\left[X_{0}\right].
\end{eqnarray*}
Note that $\underline{\Sigma}_{(x,t),(x,t)}$ is not equal to $\mathrm{Cov}_{X_{0}\mid X_{t}=x}\left[X_{0}\right]$
since in Formula (\ref{eq:cov-formula-1}), we condition on $X_{\frac{t}{2}}=x$,
not on $X_{t}=x$. Let $\Delta_{t}(x):=m_{\frac{t}{2}}(x)-m_{t}(x)=t\left[\frac{1}{2}s_{\frac{t}{2}}(x)-s_{t}(x)\right].$
Then
\begin{eqnarray*}
\underline{\Sigma}_{(x,t),(x,t)} & = & \mathbb{E}\left[(X_{0}-m_{\frac{t}{2}}(x)+\Delta_{t}(x))(X_{0}-m_{\frac{t}{2}}(x)+\Delta_{t}(x)))^{T}\frac{e^{-\frac{\mid\mid x-X_{0}\mid\mid^{2}}{t}}}{\mathbb{E}[e^{-\frac{\mid\mid x-X_{0}\mid\mid^{2}}{t}}]}\right]\\
 & = & \mathbb{E}\left[(X_{0}-m_{\frac{t}{2}}(x))(X_{0}-m_{\frac{t}{2}}(x))^{T}\frac{e^{-\frac{\mid\mid x-X_{0}\mid\mid^{2}}{t}}}{\mathbb{E}[e^{-\frac{\mid\mid x-X_{0}\mid\mid^{2}}{t}}]}\right]+\Delta_{t}(x)\Delta_{t}(x)^{T}.
\end{eqnarray*}
Hence $\underline{\Sigma}_{(x,t),(x,t)}=\mathrm{Cov}_{X_{0}\mid X_{\frac{t}{2}}=x}\left[X_{0}\right]+\Delta_{t}(x)\Delta_{t}(x)^{T}$.
This leads to the following relation between $\underline{\Sigma}$
and the Jacobian of the score: 
\begin{equation*}
\nabla s_{t}(x)=-\frac{\mathrm{Id}_{d}}{t}+\frac{1}{t^{2}}\underline{\Sigma}_{(x,2t),(x,2t)}-\frac{1}{t^{2}}\Delta_{2t}(x)\Delta_{2t}(x)^{T}.
\end{equation*}

In \cite{ventura2025manifolds, achilli2024losingdimensionsgeometricmemorization},
it is shown that the singular values of the Jacobian $\nabla s_{t}(x)$
reflect the local geometry of the data manifold. In particular, small
singular values correspond to tangent directions, whereas large values
correspond to directions orthogonal to the data manifold. 

In this work, we show that the eigenvalues of the covariance $\Sigma_{(x,t),(x,t)}$
also encodes the local geometric information of the data manifold:
small eigenvalues correspond to orthogonal manifold, whereas large
ones correspond to the tangent directions. This is natural since noise
in the data sampling mostly occurs along the manifold, with little
intensity in the orthogonal directions.

\subsection{Proof of Theorem \ref{thm: bounds bias-variance from CLT}}\label{sec:proof thm bias-variance}

\begin{proof}[Proof of Theorem \ref{thm: bounds bias-variance from CLT}]
    (i)
    Fubini's Theorem applies to show that
    \begin{align*}
        \frac{1}{N}\EE\left[\Vert (K_{h_N}\star G)(t_N,x)\Vert^2\right]
        &= \frac{1}{N}\EE\left[\int\!\!\!\int K_{h_N}(x-y)K_{h_N}(x-y')G(t_N,y)^{\mathrm{T}}G(t_N,y')\mathrm{d}y\mathrm{d}y'\right]\\
        &= \frac{1}{N}\int\!\!\! \int\frac{1}{(2\pi {h_N})^d}e^{-\frac{1}{2h_N}(\Vert x-y\Vert^2 + \Vert x-y'\Vert^2)} \mathrm{tr}\left(\Sigma_{(t_N,y),(t_N,y')}\right)\mathrm{d}y\mathrm{d}y'.
    \end{align*}
    Theorem \ref{thm: covariance asymptotics t to 0}(ii) provides the asymptotic behavior of $\Sigma_{(t_N,y),(t_N,y')}$ as $N\to\infty$, and we showed below its proof in Section \ref{sec: proof of thm covariance asymptotics} that it is uniform in $y,y'\in\RR^d$.
    In particular, the Dominated Convergence Theorem (up to rescaling by the asymptotic behavior at first order) shows that
    \begin{align*}
    \frac{1}{N}\EE\left[\Vert (K_{h_N}\star G)(t_N,x)\Vert^2\right]
    &\underset{N\to \infty}{\sim} \frac{t_N}{(2\pi t_N)^{\frac{k}{2}}}\frac{1}{N}\int\!\!\! \int\frac{1}{(2\pi h_N)^d}e^{-\frac{1}{2h_N}(\Vert x-y\Vert^2 + \Vert x-y'\Vert^2)} e^{-\frac{\Vert\pi(y)-\pi(y')\Vert^2}{4t_N}}\\
    &\hspace{1cm}\times\frac{p_*\big(\frac{\pi(y)+\pi(y')}{2}\big)}{p_*(\pi(y))p_*(\pi(y'))}\Big(k - \frac{\Vert \pi(y)-\pi(y')\Vert^2}{4t_N}\Big)\mathrm{d}y\mathrm{d}y'.
\end{align*}
To derive an upper bound, we can drop the second term inside the parenthesis. (In fact, keeping track of it yields a term that becomes negligible.)
We write $\lesssim$ for an asymptotic relation that holds for an upper bound of the left-hand side.
Integrating over the orthogonal space of $\mathcal{M}$, we obtain
\begin{align*}
    \frac{1}{N}\EE\left[\Vert (K_{h_N}\star G)(t_N,x)\Vert^2\right]
    &\lesssim\frac{1}{N}\frac{kt_N}{(2\pi t_N)^{k/2}}\int\!\!\! \int\frac{1}{(2\pi h_N)^{k}}\frac{p_*\left(\frac{\pi(y)+\pi(y')}{2}\right)}{p_*(\pi(y))p_*(\pi(y'))}\\
    &\hspace{1cm}\times e^{-\frac{1}{2h_N}(\Vert \pi(x)-\pi(y)\Vert^2 + \Vert \pi(x)-\pi(y')\Vert^2)} e^{-\frac{\Vert\pi(y)-\pi(y')\Vert^2}{4t_N}}\mathrm{d}y\mathrm{d}y'.
\end{align*}
We now identify the quadratic form in the exponentials.
We have
\begin{align*}
    \frac{1}{2h_N}&(\Vert \pi(x)-\pi(y)\Vert^2 + \Vert \pi(x)-\pi(y')\Vert^2) + \frac{\Vert\pi(y)-\pi(y')\Vert^2}{4t_N}\\
    &= \Vert\pi(x)\Vert^2\frac{1}{h_N} + (\Vert\pi(y)\Vert^2 + \Vert\pi(y')\Vert^2)\left(\frac{1}{2h_N} + \frac{1}{4t_N}\right) - \frac{1}{h_N}\langle\pi(x),\pi(y)+\pi(y')\rangle\\
    &\hspace{9cm}- \frac{1}{2t_N}\langle\pi(y),\pi(y')\rangle\\
    &= \frac{1}{2}(\pi(y)-\mu,\pi(y')-\mu)Q(\pi(y)-\mu,\pi(y')-\mu)^{\mathrm{T}},
\end{align*}
for some $Q,\mu$ to identify, where the subtracted $\mu$ is the same by symmetry of the expression in $\pi(y),\pi(y')$).
From the above, we see that the diagonal terms of $Q$ are $\frac{1}{h_N}+\frac{1}{2t_N}$.
The non-diagonal terms $Q_{j,j+k}=Q_{j+k,j}$ for $j\in\{1,\ldots,k\}$ (i.e. between $\pi(y),\pi(y')$) are $-\frac{1}{2t_N}$. The other terms are null.
We can now identify $\mu$,
\begin{align*}
    \frac{1}{2}(\mu,\mu)&Q(\mu,\mu)^{\mathrm{T}} - (\mu,\mu)Q(\pi(y),\pi(y'))^{\mathrm{T}}\\
    &= \Vert\mu\Vert^2\left(\frac{1}{h_N}+\frac{1}{2t_N}-\frac{1}{2t_N}\right) -  \langle\mu,\pi(y) + \pi(y')\rangle  \left(\frac{1}{h_N}+\frac{1}{2t_N} - \frac{1}{2t_N}\right)\\
    &= \frac{1}{h_N}\left(\Vert\mu\Vert^2 - \langle\mu,\pi(y) + \pi(y')\rangle\right).
\end{align*}
Hence, we have $\mu=\pi(x)$.
We thus get an expression in the exponential of the form
\begin{align*}
    (\pi(y)-\pi(x),\pi(y')-\pi(x))Q(\pi(y)-\pi(x),\pi(y')-\pi(x))^{\mathrm{T}}.
\end{align*}
In view of the previous calculations, we thus have that
\begin{align}\label{eq: order variance with det Q}
    \frac{1}{N}\EE\left[\Vert (K_h\star G)(t_N,x)\Vert^2\right]
    &\lesssim
    \frac{1}{N}\frac{kt_N}{(2\pi t_N)^{k/2}}\int\!\!\! \int\frac{1}{(2\pi h_N)^{k}} \frac{p_*\left(\frac{\pi(y)+\pi(y')}{2}\right)}{p_*(\pi(y))p_*(\pi(y'))}\nonumber\\
    &\hspace{1cm}\times e^{-\frac{1}{2}(\pi(y)-\pi(x),\pi(y')-\pi(x))Q(\pi(y)-\pi(x),\pi(y')-\pi(x))^{\mathrm{T}}}\mathrm{d}y\mathrm{d}y'\nonumber\\
    &\lesssim\frac{1}{N}\frac{kt}{(2\pi t_N)^{k/2}}\frac{1}{h_N^k}E(t_N,h_N,x)\mathrm{det}(Q^{-1})^{1/2},
\end{align}
where
\begin{align*}
    E(t_N,h_N,x)
    &:= \EE_{(Z,Z')\sim\mathcal{N}((\pi(x),\pi(x)),Q^{-1})}\Big[\frac{p_*\big(\frac{Z+Z'}{2}\big)}{p_*(Z)p_*(Z')}\Big].
\end{align*}
We now compute the determinant of $Q$.
Firstly, we note that
\begin{align*}
    Q = \begin{pmatrix}
        (a+b)\mathrm{Id}_k & -b\mathrm{Id}_k\\
        -b\mathrm{Id}_k & (a+b)\mathrm{Id}_k
    \end{pmatrix},
\end{align*}
where $a=\frac{1}{h_N}$ and $b=\frac{1}{2t_N}$.
For block matrix of this form, we have $\mathrm{det}\begin{pmatrix}
    A & B\\
    B & A
\end{pmatrix}= \mathrm{det}(A-B)\mathrm{det}(A+B)$, see Exercise 5.38 in \cite{abadir2005matrix}.
The determinant being the product of the eigenvalues, we deduce that
\begin{align*}
    \mathrm{det}\left((a+b)\mathrm{Id}_k -b\mathrm{Id}_k\right)
    &= a^k,
\end{align*}
and similarly,
\begin{align*}
    \mathrm{det}\left((a+b)\mathrm{Id}_k + b\mathrm{Id}_k\right)
    &= (a+2b)^k.
\end{align*}
We thus have that
\begin{align}\label{eq: determinant Q}
    \mathrm{det}(Q)
    &=\frac{1}{h_N^{k}}\times\left(\frac{1}{h_N}+\frac{1}{t_N}\right)^{k}.
\end{align}
For $h_N\gg t_N$, the asymptotic behavior of the second term in the determinant above is therefore governed by the $1/t_N$ term, that is,
\begin{align*}
    \mathrm{det}(Q)\underset{N\to\infty}{\sim} \frac{1}{h_N^kt_N^{k}}.
\end{align*}
(The case $h_N\geq t_N$ is similar up to a constant factor, hence we work with $h_N\gg t_N$ below.)

It turns out that $Q^{-1}$ can be explicitly computed. One can check by multiplying it with $Q$ that
\begin{align*}
    Q^{-1} = \begin{pmatrix}
        \frac{h_N(2t_N+h_N)}{t_N+h_N}\mathrm{Id}_k & \frac{h_N^2}{t_N+h_N}\mathrm{Id}_k\\
        \frac{h_N^2}{t_N+h_N}\mathrm{Id}_k & \frac{h_N(2t_N+h_N)}{t_N+h_N}\mathrm{Id}_k
    \end{pmatrix}.
\end{align*}
In particular, we get $\mathrm{tr}(Q^{-1})\sim 4kh_N\to 0$ as $N\to\infty$, and we deduce that
\begin{align*}
    E(t_N,h_N,x)
    &= \EE_{(Z,Z')\sim\mathcal{N}((\pi(x),\pi(x)),Q^{-1})}\Big[\frac{p_*\big(\frac{Z+Z'}{2}\big)}{p_*(Z)p_*(Z')}\Big]\underset{N\to\infty}{\longrightarrow}\frac{1}{p_*(\pi(x))}.
\end{align*}

Coming back to \eqref{eq: order variance with det Q}, we obtain for $h_N\gg t_N$ that, as $N\to\infty$,
\begin{align}\label{eq: variance convolved GP}
    \frac{1}{N}\EE\left[\Vert (K_{h_N}\star G)(t_N,x)\Vert^2\right]
    &\lesssim \frac{h_N^{k/2}t_N^{1+k/2}}{Nh_N^kt_N^{k/2}}\times\frac{kE(t_N,h_N,x)}{(2\pi)^{k/2}}\nonumber\\
    &\lesssim \frac{t_N}{Nh_N^{k/2}}\times\frac{1}{(2\pi)^{k/2}}\frac{k}{p_*(\pi(x))},
\end{align}
which proves the claim.

(ii)
For each $j\in\{1,\ldots,d\}$, we compute
\begin{align*}
    \widetilde{m}_{t_N}(x)_j-m_{t_N}(x)_j
    &=\int_{\RR^d}K_{h_N}(y)\left(m_{t_N}(x-y)_j-m_{t_N}(x)_j\right)\mathrm{d}y\\
    &=\frac{1}{\sqrt{h_N}}\int_{\RR^d}K_1(y/\sqrt{h_N})\left(m_{t_N}(x-y)_j-m_{t_N}(x)_j\right)\mathrm{d}y\\
    &=\int_{\RR^d}K_1(u)\left(m_{t_N}(x-u\sqrt{h_N})_j-m_{t_N}(x)_j\right)\mathrm{d}u.
\end{align*}
%Let $\mathcal{H}_{(m_{t_N})_j}(y)$ be the Hessian matrix of $(m_{t_N})_j$ at $y$.
Taylor's Theorem yields
\begin{align}\label{eq: taylor thm for bias}
    \widetilde{m}_{t_N}(x)_j-m_{t_N}(x)_j
    &=\int_{\RR^d}K_1(u)\nabla_xm_{t_N}(x-\lambda_u u\sqrt{h_N})_j\cdot (-\sqrt{h_N}u)\mathrm{d}u\nonumber\\
    &=\int_{\RR^d}K_1(u)\big(\nabla_xm_{t_N}(x-\lambda_u u\sqrt{h_N})_j-\nabla_xm_{t_N}(x)_j\big)\cdot (-\sqrt{h_N}u)\mathrm{d}u,
    %&=\int_{\RR^d}K_1(u)\left(-(\nabla_xm_{t_N}(x)_j)^{\mathrm{T}} \sqrt{h_N}u + \frac{h_N}{2}(u-\lambda\sqrt{h_N}u)^{\mathrm{T}}\mathcal{H}_{(m_{t_N})_j}(x)(x-\lambda \sqrt{h_N}u)\right)\mathrm{d}u,
\end{align}
for some $\lambda_u\in[0,1]$, where we used that the first moment of $K_1$ is $0$.
On the other hand, for all $x,x'\in\RR^d$, we have
\begin{align*}
    &\Vert\nabla_x m_{t_N}(x)_j - \nabla_x m_{t_N}(x')_j\Vert\\
    &\hspace{0.3cm}\leq \Vert\nabla_x m_{t_N}(x)_j - \nabla_x m_{0}(x)_j\Vert + \Vert\nabla_x m_{0}(x)_j - \nabla_x m_{0}(x')_j\Vert + \Vert\nabla_x m_{0}(x')_j - \nabla_x m_{t_N}(x')_j\Vert.
\end{align*}
The middle term is equal to $\Vert\nabla\pi(x)_j-\nabla\pi(x')_j\Vert\leq \Vert x-x'\Vert$ since $\pi(\cdot)$ is the orthogonal projection on the linear manifold $\mathcal{M}$.
Next, we write
\begin{align*}
    \nabla_x m_{t_N}(x)_j
    &= \nabla_x\frac{\int_{\mathcal{M}}z_je^{-\frac{\Vert z-x\Vert^2}{2t_N}}p_*(z)\mathrm{d}z}{\int_{\mathcal{M}}e^{-\frac{\Vert z-x\Vert^2}{2t}}p_*(z)\mathrm{d}z}\\
    &= \frac{\int_{\mathcal{M}}(z_j-m_{t_N}(x)_j)(z-x)e^{-\frac{\Vert z-x\Vert^2}{2t_N}}p_*(z)\mathrm{d}z}{\int_{\mathcal{M}}e^{-\frac{\Vert z-x\Vert^2}{2t_N}}p_*(z)\mathrm{d}z}\\
    &= \frac{\int_{\mathcal{M}}(z_j-m_{t_N}(x)_j)(z-\pi(x))e^{-\frac{\Vert z-\pi(x)\Vert^2}{2t_N}}p_*(z)\mathrm{d}z}{\int_{\mathcal{M}}e^{-\frac{\Vert z-\pi(x)\Vert^2}{2t_N}}p_*(z)\mathrm{d}z}.
\end{align*}
Laplace's Method shows that $\nabla_x m_{t_N}(x) = P_{\mathcal{M}} + O(t_N)$ as $N\to\infty$, uniformly in $x\in\RR^d$ (as shown at the end of Section \ref{sec: proof of thm covariance asymptotics}).
Hence, we have that $\Vert\nabla_x m_{t_N}(x)_j - \nabla_x m_{0}(x)_j\Vert = O(t_N)$, and then 
\begin{align*}
    \Vert\nabla_x m_{t_N}(x)_j - \nabla_x m_{t_N}(x')_j\Vert
    &\leq \Vert x - x'\Vert + O(t_N).
\end{align*}
Coming back to \eqref{eq: taylor thm for bias}, we have obtain
\begin{align*}
    \vert\widetilde{m}_{t_N}(x)_j-m_{t_N}(x)_j\vert
    &\leq\int_{\RR^d}K_1(u)\big(\lambda_u \Vert u\Vert \sqrt{h_N} + O(t_N)\big)\sqrt{h_N}\Vert u\Vert\mathrm{d}u\\
    &\leq h_N\int_{\RR^d}K_1(u)\Vert u\Vert^2 + o(h_N),
\end{align*}
and we deduce that
\begin{align*}
    \Vert \widetilde{m}_{t_N}(x)-m_{t_N}(x)\Vert^2
    &= d \Big(h_N\int_{\RR^d}K_1(u)\Vert u\Vert^2 + o(h_N)\Big)^2\\
    &= d h_N^2\EE[\chi^2]^2 + o(h_N),
\end{align*}
where $\chi$ follows a chi-distribution with $d$ degrees of freedom, so that $\EE[\chi^2] = d$.
This shows that $\Vert \widetilde{m}_{t_N}(x)-m_{t_N}(x)\Vert^2\lesssim d^3 h_N^2$.

To obtain the bound with $h_N$ instead of $h_N^2$ (which is useful only when $h_N>1$), recall the equation above \eqref{eq: taylor thm for bias}
\begin{align*}
    \widetilde{m}_{t_N}(x)_j-m_{t_N}(x)_j
    &=\int_{\RR^d}K_1(u)\left(m_{t_N}(x-u\sqrt{h_N})_j-m_{t_N}(x)_j\right)\mathrm{d}u.
\end{align*}
We write
\begin{align*}
    \Vert m_{t_N}(y)-m_{t_N}(y')\Vert
    &\leq \Vert m_{t_N}(y)-m_{0}(y)\Vert + \Vert m_{0}(y)-m_{0}(y')\Vert + \Vert m_{0}(y')-m_{t_N}(y')\Vert.
\end{align*}
Following the same argument as before yields the claim, we thus omit the details for conciseness and conclude the proof.
\end{proof}

\subsection{Proof of Theorem \ref{thm: KL regimes}}\label{sec: proof of thm KL}

\begin{proof}[Proof of Theorem \ref{thm: KL regimes}]
    We use \eqref{eq:KLinequality} to write:
    \begin{enumerate}[label = \textbullet]
    \item \textbf{With empirical score:}
    by Fubini's theorem and Theorem \ref{thm: covariance asymptotics t to 0},
    \begin{align*}
        \EE_{D}\Big[D_{\mathrm{KL}}(p_{t_N}\Vert q_{t_N}^N)\Big]
        &\leq \int_{t_N}^{T}\EE_{x_t\sim p_t}\left[\EE_{D}\left[\Vert s_t(x_t)-s_t^N(x_t)\Vert^2\right]\right]\mathrm{d}t + D_{\mathrm{KL}}(p_{T}\Vert \mathcal{N}(0,T\mathrm{Id}_d))\\
        &\leq O\bigg(\frac{1}{N}\int_{t_N}^{T}\frac{1}{t^{1+\frac{k}{2}}}\mathrm{d}t\bigg) + D_{\mathrm{KL}}(p_{T}\Vert \mathcal{N}(0,T\mathrm{Id}_d))\\
        &\leq O\bigg(\frac{1}{Nt_N^{\frac{k}{2}}}\bigg) + D_{\mathrm{KL}}(p_{T}\Vert \mathcal{N}(0,T\mathrm{Id}_d)).
    \end{align*}
    \item \textbf{With mollified score:}
    The same reasoning as above with Theorem \ref{thm: bounds bias-variance from CLT} shows that
    \begin{align*}
        \EE\left[D_{\mathrm{KL}}(p_{t_N}\Vert \tilde{q}_{t_N}^N)\right]
        &\leq O\bigg(\frac{1}{Nh_N^{k/2}}\int_{t_N}^T\frac{1}{t}\mathrm{d}t + \frac{h_N^2}{t_N}\bigg) + D_{\mathrm{KL}}(p_{T}\Vert \mathcal{N}(0,T\mathrm{Id}_d))\\
        &\leq O\bigg(\frac{\log(1/t_N)}{Nh_N^{k/2}} + \frac{h_N^2}{t_N}\bigg) + D_{\mathrm{KL}}(p_{T}\Vert \mathcal{N}(0,T\mathrm{Id}_d)),
    \end{align*}
    which proves the claim, up to a non-important $\log 1/t_N$ factor.
\end{enumerate}
\end{proof}

\subsection{Connection with change of time}
Fix a dataset $\{x_{1},\ldots,x_{N}\}$ and let $X\sim\mathcal{N}(0,\sigma^{2}\mathrm{Id}_{d})$.
Consider the case where the random variable $\frac{1}{N}\sum_{i=1}^{N}e^{-\frac{\left\Vert x+X-x_{i}\right\Vert ^{2}}{2t}}$
has low variance and can be approximated by its expectation. Then
the mollified estimator $\tilde{m}_{t}^{N}(x)=\mathbb{E}_{X}\left[m_{t}^{N}(x+X)\right]$
 can be approximated by 
\begin{equation*}
\tilde{m}_{t}^{N}(x)\simeq\frac{\mathbb{E}_{X}\left[\sum_{i=1}^{N}x_{i}e^{-\frac{\left\Vert x+X-x_{i}\right\Vert ^{2}}{2t}}\right]}{\mathbb{E}_{X}\left[\sum_{i=1}^{N}e^{-\frac{\left\Vert x+X-x_{i}\right\Vert ^{2}}{2t}}\right]}.
\end{equation*}
We now compute 
\begin{equation*}
\mathbb{E}_{X}\left[e^{-\frac{\left\Vert x+X-x_{i}\right\Vert ^{2}}{2t}}\right]=\frac{1}{(2\pi\sigma^{2})^{\frac{d}{2}}}\int_{\mathbb{R}^{d}}e^{-\frac{\left\Vert x+y-x_{i}\right\Vert ^{2}}{2t}-\frac{\left\Vert y\right\Vert ^{2}}{2\sigma^{2}}}dy.
\end{equation*}
 Completing the square and using Gaussian integrals yields 
\begin{equation*}
\mathbb{E}_{X}\left[e^{-\frac{\left\Vert x+X-x_{i}\right\Vert ^{2}}{2t}}\right]=\left[\frac{t}{\sigma^{2}+t}\right]^{\frac{d}{2}}e^{-\frac{\|x_{i}-x\|^{2}}{2(\sigma^{2}+t)}}.
\end{equation*}
Hence, we obtain the approximation of the mollified score: 
\begin{equation*}
\tilde{m}_{t}^{N}(x)\simeq\frac{\sum_{i=1}^{N}x_{i}e^{-\frac{1}{2(\sigma^{2}+t)}\|x_{i}-x\|^{2}}}{\sum_{i=1}^{N}e^{-\frac{1}{2(\sigma^{2}+t)}\|x_{i}-x\|^{2}}}=m_{t+\sigma^{2}}^{N}(x).
\end{equation*}
This shows that, when the denominator concentrates (i.e. has low variance,
possibly when $\sigma^{2}$ is small enough), mollifying the empirical score
by the Gaussian kernel $\mathcal{G}_{\sigma^{2}}$ is approximately
equivalent to considering the estimator at a larger time $t+\sigma^{2}$.
This reveals a connection between mollification and time change. 

This raises the following question: does time discretization help generalization
and prevent memorization in generative models? Indeed, during a time
step $t\in[t_{i},t_{i+1}]$, discretized sampling equation uses the
estimated score $s_{T-t_{i}}$ instead of $s_{T-t}$, effectively
evaluating the score at a larger time, hence using possibly a more
regularized estimator.

\subsection{Connection to neural networks}
\label{sec:connection-to-NN}
We propose the following heuristic picture of what might happen in the Neural Tangent Kernel (NTK) regime when the dataset is large, and why a convolution of the empirical score could naturally appear.

In the NTK regime, a neural network behaves like a kernel method: the model is mostly linear in the (recentered) parameters, with features given by $\nabla_\theta f_\theta$. At the end of training in this regime, we obtain a kernel regression with the NTK defined by $\nabla_\theta f_\theta(t,x).\nabla_\theta f_\theta(s,y)$. 

When a small $\ell_2$ regularization is applied to the parameters, assuming that the NTK regime remains valid with this small regularization, and with sufficient data points, the trained model should approximate the kernel ridge regression on the dataset $((t_i,x_i), m_{t_i}(x_i))$. 

Under suitable conditions on the kernel (e.g. existence of a Mercer decomposition), the kernel ridge regression solution can itself be interpreted as a convolution with the so-called equivalent kernel, as described in Sections 2.6 and 7.1 of \cite{rasmussen}. This provides, heuristically, a natural connection between neural networks trained in the NTK regime with regularization, and smoothing via convolution.

This interpretation is, of course, heuristic. Still, we believe it offers some valuable intuition about how regularization, large dataset sizes, and NTK regime interact to create a regularization which could be a convolution in space and time.  

Note that equivalent kernels are generally not positive and do not integrate to one. In this paper, we mostly consider smoothing by Gaussian kernels  (which are positive and integrate to one), but one could consider other, potentially better, non-positive kernels that do not integrate to one. This observation is also supported by the LDE-KDE framework, in which the kernel which comes from the regularisation operates in log-density space. In this framework, the kernel does not need to be positive and no condition on its integral is required. A simple, yet interesting example of kernel to investigate could be $(1+\alpha) \mathcal{G}_{\sigma_1^2}-\alpha \mathcal{G}_{\sigma_2^2}$ where $\sigma_2\ll \sigma_1$ so that $\mathcal{G}_{\sigma_2^2}(x,y)\sim \delta_0(x,y)$. As in classifier-free guidance \cite{luo2022understandingdiffusionmodelsunified,ho2022classifierfreediffusionguidance}), this kernel penalize regions lying too close to the training points. 
%https://arxiv.org/pdf/2305.07241
% https://proceedings.mlr.press/v130/chen21e/chen21e.pdf et citations

%\francois{\textbf{To make the link between NN and convolution of the equivalent kernel}, we can formally assume: 1. the NN is infinitely wide in NTK regime (i.e. kernel method with kernel $\Theta((t,x),(t',x'))$). 2. the number of data then tends to infinity.
%Let $s_*$ be the ground truth, then the solution to the regression pb in the RKHS is (sthg like)
%\begin{align*}
 %   s(x,t)
  %  =\frac{1}{N^2}\sum_{i,j=1}^N \Theta((t,x),(t_i,x_i))\overline{\Theta}^{-1}((t_i,x_i),(t_j,x_j))s_*(x_j,t_j),
%\end{align*}
%where $\overline{\Theta}$ is the Gram matrix of the data (i.e. the kernel $\Theta$ at the data points).
%Now let $N$ to infinity and we hope to have a convergence (to check)
%\begin{align*}
 %   s(x,t)
  %  &=\int\!\!\int K((t,x),(\tau,y))K^{-1}((\tau,y),(\sigma,z))s_*(z,\sigma)\mathrm{d}y\mathrm{d}z\mathrm{d}\tau\mathrm{d}\sigma\\
   % &=(H*s_*)(t,x),
%\end{align*}
%for $H$ a kernel that should look like the equivalent kernel of $\Theta$.
%}

\subsection{The spectral point of view}

So far, our analysis relies on approximating the variance and the bias of the estimator $\tilde{s}_t^N$ using the specific covariance structure of the noise following the CLT \eqref{eq: CLT for m_N}.
In this section, we present a different heuristic approach based on the spectral decomposition of the heat semigroup.
We make several simplifying assumptions and do not aim at the greatest generality.
In particular, as opposed to the rest of the paper, we assume in this section that the measure $p_*$ has full support in the ambient space.
We stress that \textbf{the approach below is heuristic} while the other results of this work are rigorously established.

\paragraph{Brownian motion diffusion setup}
Let $H_d:=[-1,1]^d$ be the $d$-dimensional hypercube and suppose $p_*$ has full support in $H_d$.
Consider the heat equation in $H_d$ with zero von Neumann boundary condition
\begin{align*}
    \begin{cases}
        \partial_t u(t,x) = \Delta u(t,x),
        &\forall x\in \overset{\circ}{H}_d,\ \forall t\geq 0,\\
        \nabla_x u(t,x) = 0,
        &\forall x\in \partial H_d,\ \forall t\geq 0,
    \end{cases}
\end{align*}
where $\overset{\circ}{H}_d$ denotes the interior of $H_d$ and $\partial H_d$ its boundary.
For all $k\in\NN$, let $f_{k}(x):=\cos(\pi k x)$.
One can show that the Laplacian has eigenfunctions and respective eigenvalues given for all $\mathbf{k}\in\NN^{d}$ by
\begin{align*}
    f_{\mathbf{k}}(x)
    &= \prod_{\ell=1}^{d}f_{k_\ell}(x_\ell),\\
    \lambda_{\mathbf{k}}
    &= -\pi^2\Vert \mathbf{k}\Vert^2.
\end{align*}

We consider the diffusion starting from initial condition $p_0=p_*$ and that starting from the empirical measure $p_0^N$, with a Brownian motion in $H_d$ reflected on the boundaries as a noising process.
As usual, we denote by $p_t$ and $p_t^N$ the corresponding distributions at time $t\geq 0$.
We have the spectral decomposition 
\begin{align*}
    p_t(x)
    &=\int_{H_d}\sum_{\mathbf{k}\in\NN^d}e^{-\pi^2\Vert\mathbf{k}\Vert^2t}f_{\mathbf{k}}(x)f_{\mathbf{k}}(y)p_0(y)\mathrm{d}y,
\end{align*}
and similarly for $p_t^N$.
Let $g_{k_\ell}(x_\ell):=\sin(\pi k_\ell x_\ell)$ and note that $\partial_{x_\ell} (f_{k_\ell}(x_\ell))=-\pi k_\ell g_{k_\ell}(x_\ell)$.
For measure $\mu$ on $H_d$ and a $\mu$- integrable map $h:H_d\to\RR$, we write $\langle \mu,h\rangle:=\int_{H_d}h(y)\mu(\mathrm{d}y)$.
The score can be written as 
\begin{align*}
    s_t(x)
    &= \left(\sum_{\mathbf{k}\in\NN^d} e^{-\pi^2\Vert \mathbf{k}\Vert^2t}\frac{-\pi k_m}{p_t(x)}\frac{g_{k_m}(x_{m})}{f_{k_m}(x_m)}f_{\mathbf{k}}(x) \big\langle p_0,f_{\mathbf{k}}\big\rangle\right)_{m=1,\ldots,d}.
\end{align*}
Similarly, for the empirical score, we have
\begin{align*}
    s_t^N(x)
    &= \left(\sum_{\mathbf{k}\in\NN^d} e^{-\pi^2\Vert \mathbf{k}\Vert^2t}\frac{-\pi k_m}{p_t^N(x)}\frac{g_{k_m}(x_{m})}{f_{k_m}(x_m)}f_{\mathbf{k}}(x) \big\langle p_0^N,f_{\mathbf{k}}\big\rangle\right)_{m=1,\ldots,d}.
\end{align*}

\paragraph{Bias-variance decomposition in frequency space}

Let $x\in H_d$ and $t,h\in (0,\infty)$ be fixed.
As usual, $\tilde{s}_t$ denotes the mollified score $\mathcal{G}_h\star s_t$ and similarly for the mollified empirical score $\tilde{s}_t^N$.
Taking the expectation over the dataset $D=\{x^{i};i=1,\ldots,N\}$, akin to \eqref{eq: bias-variance decomposition} but directly on the score below, one obtains the bias-variance decomposition
\begin{align}\label{eq: frequency bias-variance decomposition}
    \EE_D\Big[\Vert \tilde{s}_t^N(x) - s_t(x)\Vert^2\Big]
    &\leq 2\underbrace{\EE_{D}\Big[\Vert \tilde{s}_t^N(x) - \tilde{s}_t(x) \Vert^2\Big]}_{v_N(t,h,x)} + 2\underbrace{\Vert \tilde{s}_t(x) - s_t(x)\Vert^2}_{b(t,h,x)}.
\end{align}
Assuming that we have a concentration $p_t^{N}\approx p_t$ and treating $p_t(y)\approx p_t(x)$ as a constant for $y$ in a neighborhood of $x$, we can write
\begin{align*}
    \tilde{s}_t(x)
    &= \bigg(\sum_{\mathbf{k}\in\NN^d}e^{-\pi^2\Vert \mathbf{k}\Vert^2 (t+h)}\frac{-\pi k_m g_{k_m}(x_m)}{p_t(x)f_{k_m}(x_m)}f_{\mathbf{k}}(x)\big\langle p_0,f_{\mathbf{k}}\big\rangle\bigg)_{m=1,\ldots,d}.
\end{align*}
The analogue formula holds for $\tilde{s}_t^{N}$ with $p_0^N$ in place of $p_0$.
One then obtains the following expressions for the variance and the bias of the estimator $\tilde{s}_t^N$:
    \begin{align*}
        v_N(t,h,x)
        &=\sum_{m=1}^d\EE_D\bigg[\bigg(\sum_{\mathbf{k}\in\NN^d}e^{-\pi^2\Vert \mathbf{k}\Vert^2 (t+h)}\frac{-\pi k_m }{p_t(x)}\frac{g_{k_m}(x_{m})}{f_{k_m}(x_m)}f_{\mathbf{k}}(x) \big\langle p_0^N-p_0, f_{\mathbf{k}}\big\rangle\bigg)^2\bigg],\\
        b(t,h,x)
        &=\sum_{m=1}^d\bigg(\sum_{\mathbf{k}\in\NN^d}e^{-\pi^2\Vert \mathbf{k}\Vert^2 t}\left(1-e^{-\pi^2\Vert\mathbf{k}\Vert^2 h}\right)\frac{-\pi k_m }{p_t(x)}\frac{g_{k_m}(x_{m})}{f_{k_m}(x_m)}f_{\mathbf{k}}(x)\big\langle p_0, f_{\mathbf{k}}\big\rangle\bigg)^2.
    \end{align*}

The control of the bias depends on the regularity of $p_0$ and can be seen from the double cut-off $e^{-\pi^2\Vert \mathbf{k}\Vert^2 t}\left(1-e^{-\pi^2\Vert\mathbf{k}\Vert^2 h}\right)\approx \mathbf{1}_{\{\pi^{-2}h^{-1} \leq \Vert k\Vert^2\leq \pi^{-2}t^{-1}\}}$, which truncates all frequencies smaller than $\pi^{-1}h^{-\frac{1}{2}}$ and larger than $\pi^{-1}t^{-\frac{1}{2}}$.
Assuming $p_0$ is smooth and has full support with density bounded away from zero entails that $\nabla \log p_t=\frac{\nabla p_t}{p_t}$ is uniformly bounded, which ensures that the bias remains finite.
We thus only focus on the variance below.

In $v_N(t,h,x)$, the regularizing effect of convolution is to truncate frequencies outside of the $\ell_2$-ball $B_d(0,\pi^{-1}(t+h)^{-\frac{1}{2}})$, since $e^{-\pi^2\Vert\mathbf{k}\Vert^2(t+h)}\approx \mathbf{1}_{\{\Vert\mathbf{k}\Vert^2\geq \pi^{-2}(t+h)^{-1}\}}$.
Using a multi-dimensional CLT, when $\Vert \mathbf{k}\Vert\leq \pi^{-1}(t+h)^{-\frac{1}{2}}$, for $N$ large enough, 
\begin{equation*}
\big\langle p_0^N-p_0, f_{\mathbf{k}}\big\rangle\ \approx \frac{1}{\sqrt{N}} \xi(f_{\mathbf{k}})
\end{equation*}
where $(\xi(f_{\mathbf{k}}))_{\Vert \mathbf{k}\Vert\leq \pi^{-1}(t+h)^{-\frac{1}{2}}}$ is a centered Gaussian variable with 
covariance $\mathbb{E}[\xi(f_{\mathbf{k}})\xi(f_{\mathbf{k'}})] = \mathrm{Cov}_{X\sim p_0}[f_{\mathbf{k}}(X), f_{\mathbf{k}'}(X)]$. Hence for $m=1,\ldots,d$, 
\begin{align*}
    &\sum_{\substack{\mathbf{k}\in\NN^d \\ \Vert \mathbf{k}\Vert\leq \pi^{-1}(t+h)^{-\frac{1}{2}}}}k_m\frac{g_{k_m}(x_{m})}{f_{k_m}(x_m)}f_{\mathbf{k}}(x) \big\langle p_0^N-p_0, f_{\mathbf{k}}\big\rangle
    \approx\frac{1}{\sqrt{N}}\sum_{\substack{\mathbf{k}\in\NN^d \\ \Vert \mathbf{k}\Vert\leq \pi^{-1}(t+h)^{-\frac{1}{2}}}}k_m\frac{g_{k_m}(x_{m})}{f_{k_m}(x_m)}f_{\mathbf{k}}(x)\xi(f_{\mathbf{k}}).
\end{align*}
For almost all $x\in H_d$ and most $\mathbf{k}$ with large norm, $f_{\mathbf{k}}$ oscillates fast and viewing $\mathbf{k}$ as a random multi-index sampled uniformly in the corresponding ball with large radius, we expect $f_{\mathbf{k}}(x)g_{k_m}(x_m)/f_{k_m}(x_m)$ to behave as independent centered random variables in $[-1,1]$, also independent from $\xi(f_{\mathbf{k}})$.
For $1\leq k_m\leq \pi^{-1}(t+h)^{-\frac{1}{2}}$, define $r(k_m):=(\pi^{-2}(t+h)^{-1} - k_m^2)^{\frac{1}{2}}$.
Since the $f_{\mathbf{k}}$s are bounded, all the covariances of interest are bounded, and thus, at least intuitively, we expect after a use of Lyapunov-CLT, on the frequencies $\mathbf{k}$ this time, that
\begin{align*}
    &\sum_{\substack{\mathbf{k}\in\NN^d \\ \Vert \mathbf{k}\Vert\leq \pi^{-1}(t+h)^{-\frac{1}{2}}}}k_m\frac{g_{k_m}(x_{m})}{f_{k_m}(x_m)}f_{\mathbf{k}}(x) \big\langle p_0^N-p_0, f_{\mathbf{k}}\big\rangle\\
    &\hspace{2cm}\approx \frac{1}{\sqrt{N}}\bigg(\sum_{k_m=1}^{\pi^{-1}(t+h)^{-\frac{1}{2}}}k_m^2\mathrm{Vol}\big(B_{d-1}(0,r(k_m)\cap\NN^{d-1})\big) \bigg)^{\frac{1}{2}}Z,
\end{align*}
where $Z\overset{d}{\sim}\mathcal{N}(0,V(x))$ with $V$ bounded.
Letting $R=\pi^{-1}(t+h)^{-\frac{1}{2}}$, the sum can be approximated by the integral
\begin{align*}
    \int_1^R u^2(R^2-u^2)^{\frac{d-1}{2}}\mathrm{d}u
    &\leq R\int_1^R u(R^2-u^2)^{\frac{d-1}{2}}\mathrm{d}u
    = \frac{R}{d+1}\Big[(R^2-u^2)^{\frac{d+1}{2}}\Big]_1^R\\
    &\underset{R\to\infty}{\sim} C R^{d+2}.
\end{align*}
 Hence, we obtain up to some multiplicative constant that does not depend on $t$ nor $x$ that
\begin{align*}
    v_N(t,h,x)
    &\approx\bigg(\sum_{\substack{\mathbf{k}\in\NN^d \\ \Vert \mathbf{k}\Vert\leq \pi^{-1}(t+h)^{-\frac{1}{2}}}}k_m\frac{g_{k_m}(x_{m})}{f_{k_m}(x_m)}f_{\mathbf{k}}(x) \big\langle p_0^N-p_0, f_{\mathbf{k}}\big\rangle\bigg)^2\\
    &\lesssim\frac{C}{N}\frac{1}{(t+h)^{1+\frac{d}{2}}}.
\end{align*}

\paragraph{Improvement of the KL bound with adaptive lengthscale}
Using \eqref{eq:KLinequality} as in Section \ref{sec: proof of thm KL} yields, for $h_N\gg t_N>0$ with $h_N\to 0$, that as $N\to\infty$, 
\begin{align*}
    \EE_{D}[D_{\mathrm{KL}}(p_{t_N}\Vert\tilde{q}_{t_N}^N)]
    &\leq O\left(\frac{1}{N h_N^{\frac{d}{2}}}\right) + D_{\mathrm{KL}}(p_{T}\Vert \mathcal{N}(0,T\mathrm{Id}_d)).
\end{align*}
We recover, up to a log factor, the statement of Theorem \ref{thm: KL regimes} (ii).
However, choosing an adaptive $h=h(t)=t^{\beta}$ for some $\beta\in(0,1)$ to construct $\tilde{s}_t^N=\mathcal{G}_{h(t)}\star s_t^N$, we obtain from the variance bound obtained just above that
\begin{align}\label{eq: bound KL spectral adaptive h}
    \EE_{D}[D_{\mathrm{KL}}(p_{t}\Vert\tilde{q}_{t}^N)]
    &\leq O\left(\frac{1}{N t^{\frac{\beta d}{2}-(1-\beta)}}\right) + D_{\mathrm{KL}}(p_{T}\Vert \mathcal{N}(0,T\mathrm{Id}_d)).
\end{align}
This improves the bound that one would obtain, even with an adaptive $h$, from the bias-variance analysis of Theorem \ref{thm: bounds bias-variance from CLT}.
Indeed, combining \eqref{eq: order variance with det Q} and \eqref{eq: determinant Q}, one obtains (with $k=d$)
\begin{align*}
    \EE\left[\Vert \tilde{s}_t^N(x)-\tilde{s}_t\Vert^2\right]
    &\lesssim \frac{C}{t^2}\frac{t}{h(t)^dt^{\frac{d}{2}}}\times h(t)^{\frac{k}{2}}\Big(\frac{1}{t^d} + \frac{1}{h(t)^d}\Big)^{-\frac{1}{2}}\\
    &\approx C\frac{1}{t^{1+\frac{\beta d}{2}}},
\end{align*}
and we deduce (considering only the variance term, as for the spectral study)
\begin{align}\label{eq: bound KL covariance adaptive h}
    \EE_{D}[D_{\mathrm{KL}}(p_{t}\Vert\tilde{q}_{t}^N)]
    &\leq O\left(\frac{1}{N t^{\frac{\beta d}{2}}}\right) + D_{\mathrm{KL}}(p_{T}\Vert \mathcal{N}(0,T\mathrm{Id}_d)).
\end{align}
Therefore, the bound \eqref{eq: bound KL spectral adaptive h} from the spectral analysis has an additional factor $t^{1-\beta}$ that mitigates the small-time explosion compared to the bound \eqref{eq: bound KL covariance adaptive h}.

This heuristic approach, based on spectral decomposition, thus suggests that the effect of regulatization could be even stronger than what is proven in Theorem \ref{thm: KL regimes}.
\newpage
\section{Numerical experiments}
\label{ap:numerical}
In this section, we provide details for the numerical experiments presented in the main part of the paper, as well as further experiments. All the codes will be made publicly available through a github repository.

\subsection{LED-KDE with other kernels }

Figure \ref{fig:LED-KDE} shows the LED-kernel density estimator, when both kernels---the one smoothing the empirical measure, and the one applied in the log-density space---are Gaussian. For numerical stability, we add $\epsilon  = 10^{-10}$ to the KDE's density before taking the logarithm. 

\begin{figure}[h!]
\centering
\includegraphics[scale=0.6]{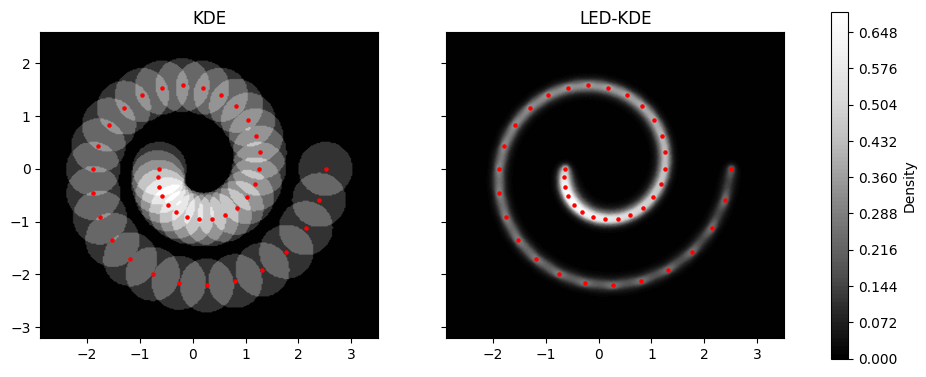}
\caption{Left: KDE with kernel $C_{0.5}$. Right: LED-KDE  $(C_{0.47}, C_{0.5}) \star p^N_0$.}
\label{fig:LED-KDE-Circle}
\end{figure}

In fact, to obtain a density estimator, we can use any kernels (positive or not, as long as things are well defined, see also Section \ref{sec:connection-to-NN}). In Figure \ref{fig:LED-KDE-Circle}, we use kernels of the form 
$$C_r = \frac{1}{\pi r^2} 1\!\!1_{\| x-y \| \leq r},$$ and plot the density $(C_{r'}, C_{r}) \star p^N_0$ with $r=0.5$ and $r'=0.47$. 

We clearly see the two distinct effects of the two kernels. The first, acting on the empirical measure, connects nearby points and reveals a structure. The second kernel, applied in log-density space, smooths and refines this structure, and spreads mass along this structure. 

\subsection{Two-dimensional Swiss-roll}

We consider the distribution $p_*$ of the random vector $(\theta * \cos(\theta), \theta*\sin(\theta))$ with $\theta \sim U([\pi,4\pi])$. The support of $p_*$ is a spiral. Our dataset then consists of $100$ i.i.d. points independently sampled from this $p_*$. 

\begin{figure}[h!]
    \centering
\includegraphics[width=0.45\textwidth]{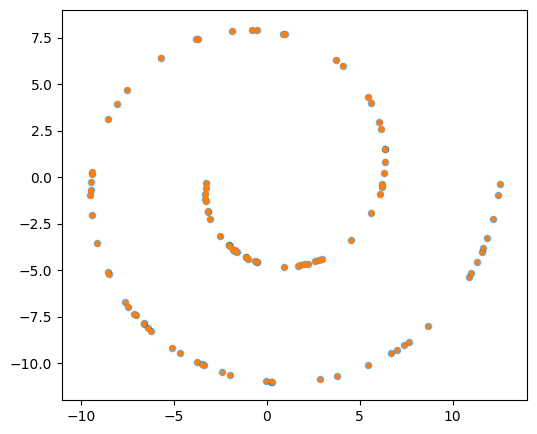}
\includegraphics[width=0.45\textwidth]{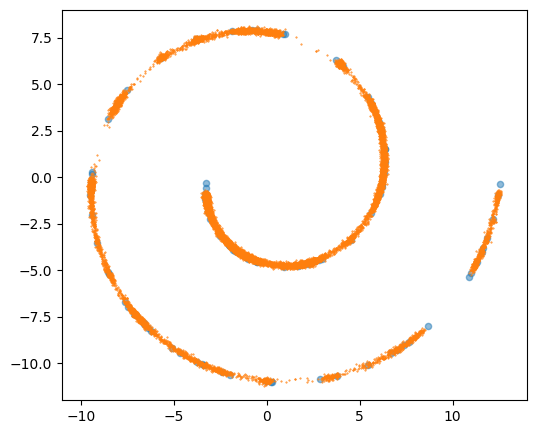}
    \caption{Generation of 10 000 points (orange), using a dataset of 100 points on the swiss-roll (blue). Left: using the empirical score. Right: using the mollified score with $h=0.75$.}
    \label{fig:SR-two}. 
\end{figure}

For all experiments, we set $\sigma = 1$, final time $T = 50$, time-step $\Delta t = 2\times 10^{-3}$, and sampling time $t_N= 2\times 10^{-3}$. We sample 10 000 points using the generative diffusion equation, up to time $T-t_N$. In Figure \ref{fig:SR-two} (left image), we plot the dataset (in blue) and the samples (in orange) obtained using the empirical score. In Figure \ref{fig:SR-two} (right image), we use the mollified score with $h=0.75$. 

\begin{figure}[h!]
    \centering
\includegraphics[width=1.\textwidth]{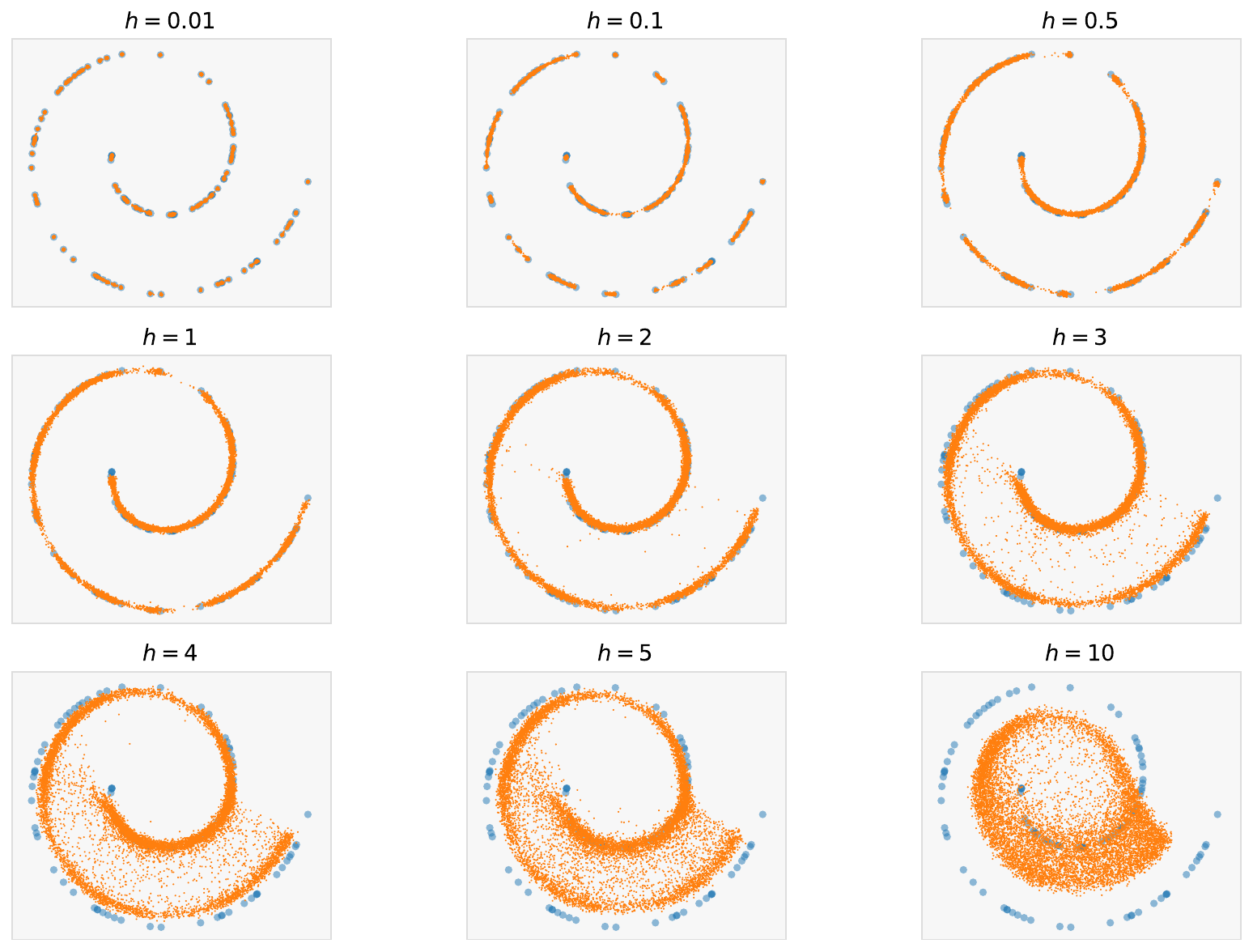}
    \caption{Generation of 10 000 points (orange), using a dataset of 100 points on the swiss-roll (blue), using various levels of regularization.}
    \label{fig:SR-comparison}
\end{figure}

Figure \ref{fig:SR-comparison} illustrates the effect of $h$: 
\begin{itemize}
\item \textbf{Small $h$}: memorization of the training points and no generation of any new points. 
\item \textbf{Moderate $h$}: memorization decreases and sampling begins to generalize. We recover a large part of the spiral. 
\item \textbf{Large $h$}: bias grows, samples appear outside the manifold, and for very large $h$ the generated distribution no longer looks like the target distribution.
\end{itemize}

\subsection{Empirical covariance}
\begin{figure}[h]
    \centering
\includegraphics[width=1.0\textwidth]{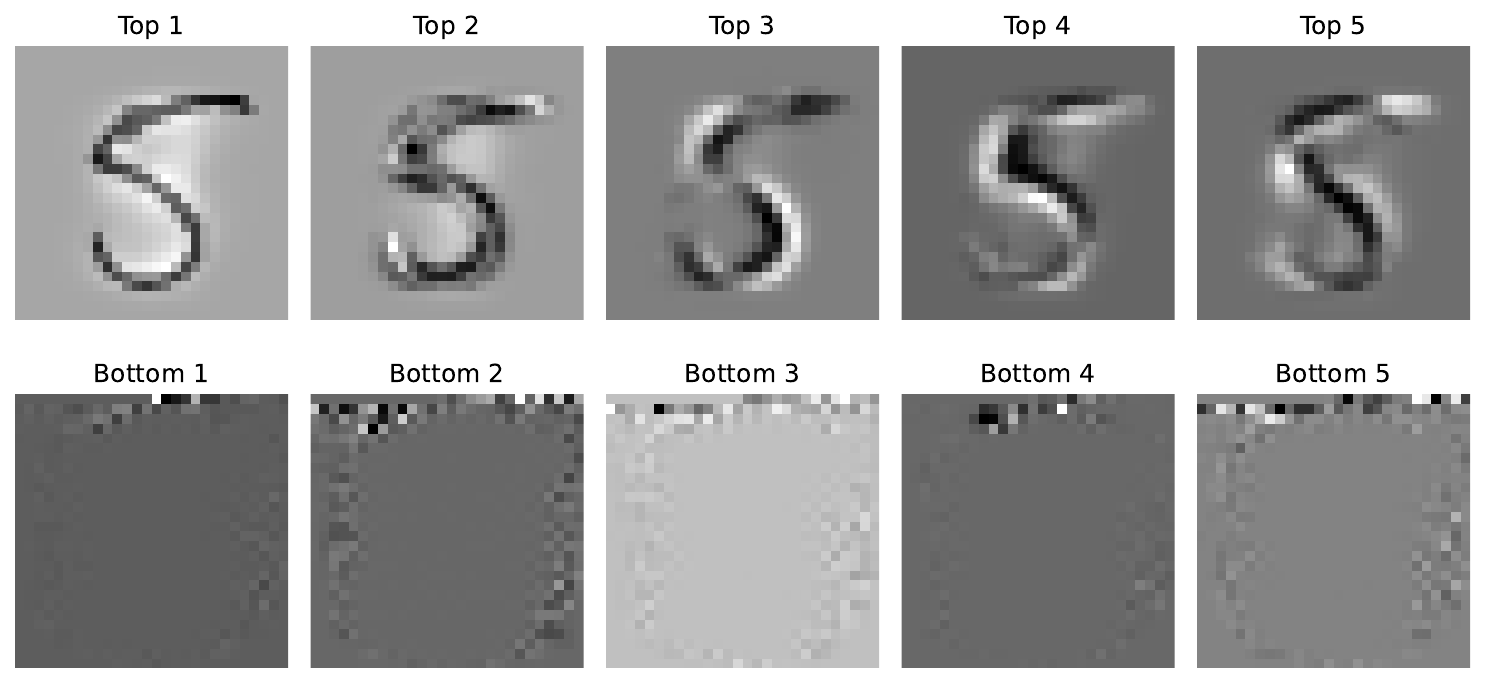}
    \caption{Eigenvectors associated with top 5 and bottom 5 eigenvalues, for the local covariance matrix at a training datapoint.}
\label{fig:MNISTcovariance}
\end{figure}

In Figure \ref{fig:MNISTcovariance}, we plot the top and bottom eigenvectors of the covariance matrix $\Sigma_{(x,t),(x,t)}$ for the MNIST dataset, similarly to Figure \ref{fig:SRcovariance}, which was for the Swiss Roll. 
One sees in Figure \ref{fig:MNISTcovariance} that locally, the first five principal eigenvectors are directions along which the image can be modified while preserving the structure of the digit $5$.
The last five eigenvectors of the matrix are locally orthogonal to the data, which can roughly speaking be seen from the fact that the center of the images, where digits typically appear, are monochromatic with no noise.

\subsection{Generalization and Effective Dataset Size}

The setup used to generate Figure \ref{fig:effectiveN} is the following: $p_*$ is a Gaussian distribution $\mathcal{N}(0,\mathrm{I}_{4\times 4})$ in $\mathbb{R}^4$. We set $
\sigma = 1$, $T=15$, and $\Delta t = t_N/10$. We approximate the score with $N=100$ samples.

We compare the KL-divergence between $\mathcal{G}_{t_N}\star p_*$ (a Gaussian distribution $\mathcal{N}(0,(1+t_N)\mathrm{I}_{4\times 4})$) and the empirical measure $q_{t_N}$, computed using the empirical score and the mollified score. %\textcolor{red}{We did not compare the KL between gtn star p* and ptN* and the KL between qtN and ptN* ? In the Figure \ref{fig:effectiveN}, we should then change the caption, and also in the legend of the y axis, there is a problem since it is written the $ln (D_Kl ( ... ))$, we should just put $ln(D_{KL}( \cdot \| p_t))$, no ? }
The empirical measures $q_{t_N}$ and $\tilde{q}_{t_N}$ are approximated by \cite{song2021scorebased}: 
\[ q_{t_N}(x) =  \exp \left(-\frac{1}{2}\int_{t_{N}}^T \nabla \cdot s^N_{t}(x_t) dt \right) q_T(x_T), \quad q_T \sim \mathcal{N}(0,T \mathrm{I}_{4\times 4}), \]
\[ \tilde{q}_{t_N}(x) =  \exp \left(-\frac{1}{2}\int_{t_{N}}^T \nabla \cdot (K\star s^N_{t})(x_t) dt \right) q_T(x_T), \quad \tilde{q}_T \sim \mathcal{N}(0,T \mathrm{I}_{4\times 4}), \]
respectively, with $dx_t=-\frac{1}{2} s_t(x_t)dt$, with $x_0 = x$.

We compute the divergences using automatic differentiation. As explained in \ref{ap:covariance_and_score}, the divergence of the empirical score---and hence of its mollified version---has a closed-form expression that can also be used directly.

%\textcolor{blue}{Maria : inside the exponent is $\exp\left(-\frac{1}{2}\int_{t_N}^T \frac{1}{2} \nabla.s_t(x_t)\right)$  or $\exp\left(-\int_{t_N}^T \frac{1}{2} \nabla.s_t(x_t)\right)$ -- i thought it was 1/2 as in dropbox not 1/4 - also song paper}\textcolor{red}{Fr: Yes this is fine the 1/2, would have say the same. }

\begin{figure}[h!]
    \centering
\includegraphics[width=1.0\textwidth]{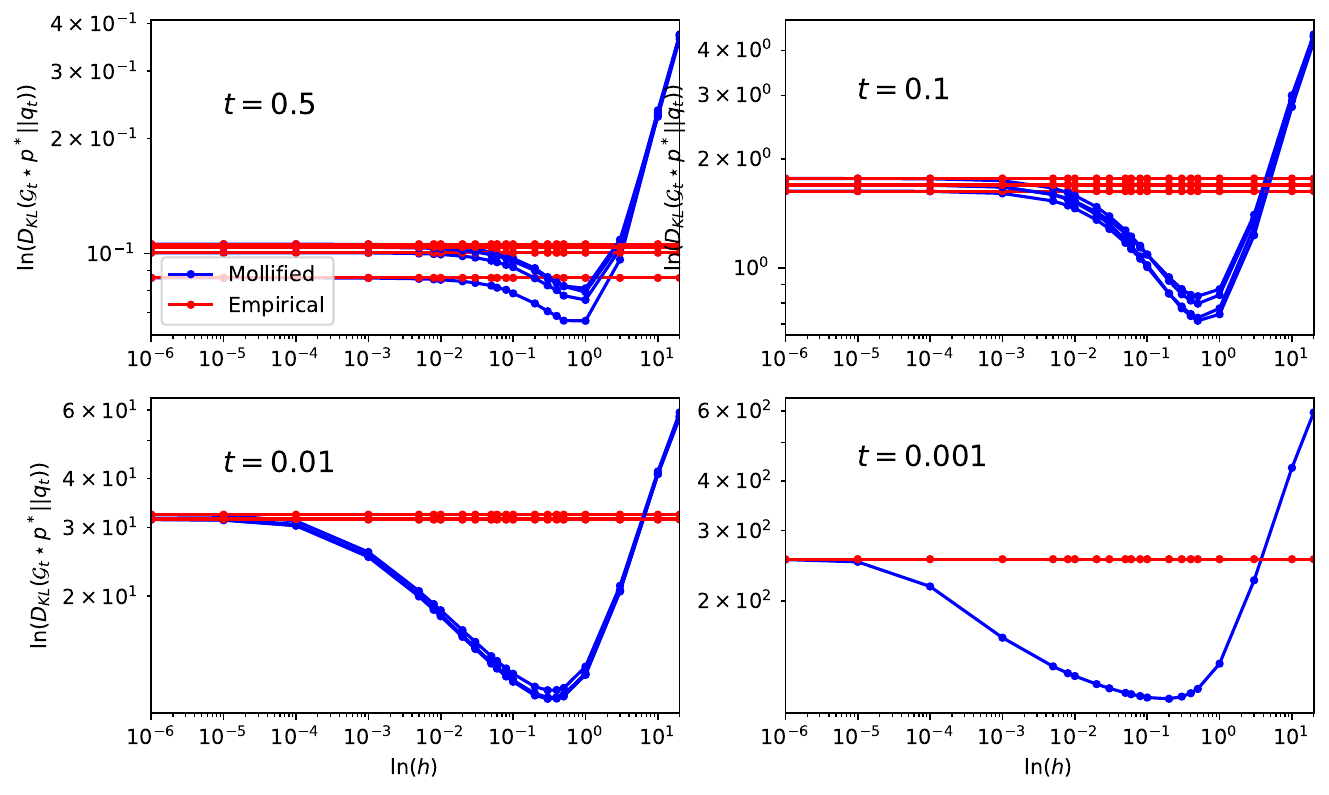}
\caption{KL-divergence between $\mathcal{G}_{t_N}\star p_*$ and the  empirical measure
generated by following the score (red)  and the KL-divergence between  $\mathcal{G}_{t_N}\star p_*$ and the empirical measure generated by following the mollified score, varying $h$  (blue). $p_*$ is multi-dimensional Gaussian ($d=4$) and $N=100$.}\label{fig:n_effec_appendix}
\end{figure}

In Figure \ref{fig:n_effec_appendix}, we show how the KL-divergence changes with respect to sampling time $t_N$ and the convolution bandwidth $h$. Numerically, for each $t_N$, we found the $h$ that yielded the lowest KL-divergence between $\mathcal{G}_{t_N}\star p_*$ and $\tilde{q}_{t_N}$, described in Table \ref{table:app_h}. The KL-divergence is approximated with $Q=500$ points. The estimated $N_{\mathrm{eff}}$ for this experiment is shown in Figure \ref{fig:effectiveN} (right).

%In Figure \ref{fig:kl_d=10}, we consider the same setup as before, with $d=10$.

\begin{table}[h!]
\centering
\begin{tabular}{|c c |} 
 \hline
 $t_N$ & $h$ \\ [0.5ex] 
 \hline\hline
  0.5 & 1.0\\
 0.1 & 0.5 \\
 0.01 & 0.3 \\
 0.001 & 0.2 \\ [1ex] 
 \hline
\end{tabular}
\caption{Optimal $h$ for each $t_N$ (numerically obtained).}
\label{table:app_h}
\end{table}

%\textcolor{red}{Fr : we need to correct all the captions with the KL-divergence (just as a reminder)}

\begin{figure}[h!]
    \centering
\includegraphics[width=1.0\textwidth]{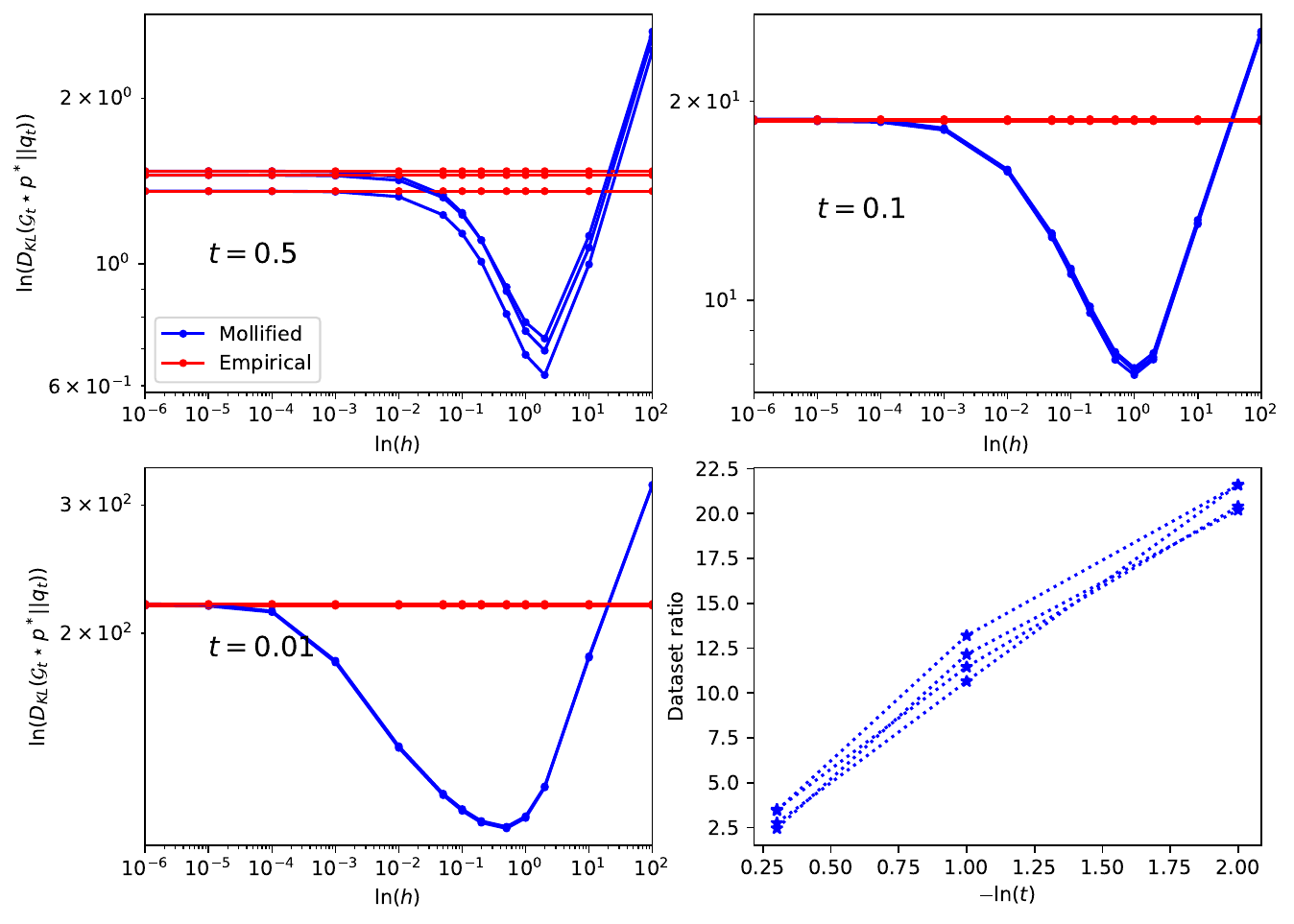}
\caption{1-3 figures: KL-divergence between $\mathcal{G}_{t_N}\star p_*$ and the  empirical measure
generated by following the score (red)  and the KL-divergence between  $\mathcal{G}_{t_N}\star p_*$ and the empirical measure generated by following the mollified score, varying $h$  (blue). 4th figure: Ratio $N_{\mathrm{eff}}/N$ at the lowest reported KL-divergence. $p_*$ is a multi-dimensional Gaussian ($d=10$) and $N=100$.}\label{fig:kl_d=10}
\end{figure}

We repeat the same experiment with $d=10$, as shown in Figure \ref{fig:kl_d=10}. We attain similar results, where for some bandwidth $h$, the KL-divergence between $\mathcal{G}_{t_N}\star p^*$ and the empirical measure $\tilde{q}_{t_N}$ computed using the mollified scores is significantly smaller than the one computed using the empirical score. Considering the empirically obtained optimal $h$, we compute the dataset ratio, between $N_{\mathrm{eff}}$ and $N$, showing for example, that for sampling time $t_N = 10^{-2}$, a dataset of size $\approx 20\times N$, when using the empirical score, is necessary to attain the same KL-divergence as when using the mollified score.

\paragraph{Hyper-sphere case.} We consider another example, where $p^*$ is a uniform distribution over a $d=4$ dimensional sphere of radius $1$. Samples from $p^*$ are generated by taking samples from a Gaussian distribution $\mathcal{N}(0,\mathrm{I}_{4\times 4})$ and dividing them by its norm.

We can write the density of $p_{t}=\mathcal{G}_{t}\star p_{*}: 
 p_{t}(x)=f_{t}(\|x\|)$ in closed form:
%\textcolor{red}{Fr : I let you add here / modify the following in the paragraph you will write on the experiments:} 
%A closed form exists for the density of $p_{t}=\mathcal{G}_{t}\star p_{*}$: 
% $p_{t}(x)=f_{t}(\|x\|)$ where 
\begin{equation*}
f_{t}(r)=\frac{1}{(2\pi t)^{\frac{d}{2}}}\frac{1}{\sqrt{\pi}}\frac{\Gamma(\frac{d}{2})}{\Gamma(\frac{d-1}{2})}e^{-\frac{r^{2}+1}{2t}}\int_{0}^{\pi}e^{\frac{r\cos\phi}{t}}(\sin\phi)^{d-2}d\phi.
\end{equation*}
This expression is obtained by using the rotational invariance of the density, thus, considering only $x=re_{1}$. We decompose $y=(\cos\phi)e_{1}+y^{\perp}$
where $y^{\perp}$ is orthogonal to $e_{1}$, and we slice the integral
according to the angle $\phi$. To estimate
the integral $\int_{0}^{\pi}e^{\frac{r\cos\phi}{t}}(\sin\phi)^{d-2}d\phi$, highly concentrated around $\pi/2$ because of the term
$(\sin\phi)^{d-2}$, we do a Monte-Carlo
method with respect to the density $\propto(\sin\phi)^{d-2}d\phi$, the law of $\phi$ when $y$ is uniform on the sphere. This
method is almost equivalent to approximating $\mathcal{G}_{t}\star p_{*}$
directly using $\mathcal{G}_{t}\star p_{*}(x)\simeq\frac{1}{(2\pi t)^{\frac{d}{2}}}\frac{1}{N}\sum_{i=1}^{N}e^{-\frac{\|x-y_{i}\|^{2}}{2t}}$
where $y_{1},\ldots,y_{N}$ are i.i.d and uniform on the sphere, the
main difference being on the fact that we impose rotational invariance
of the estimated density. 

\begin{figure}[h!]
    \centering
\includegraphics[width=1.0\textwidth]{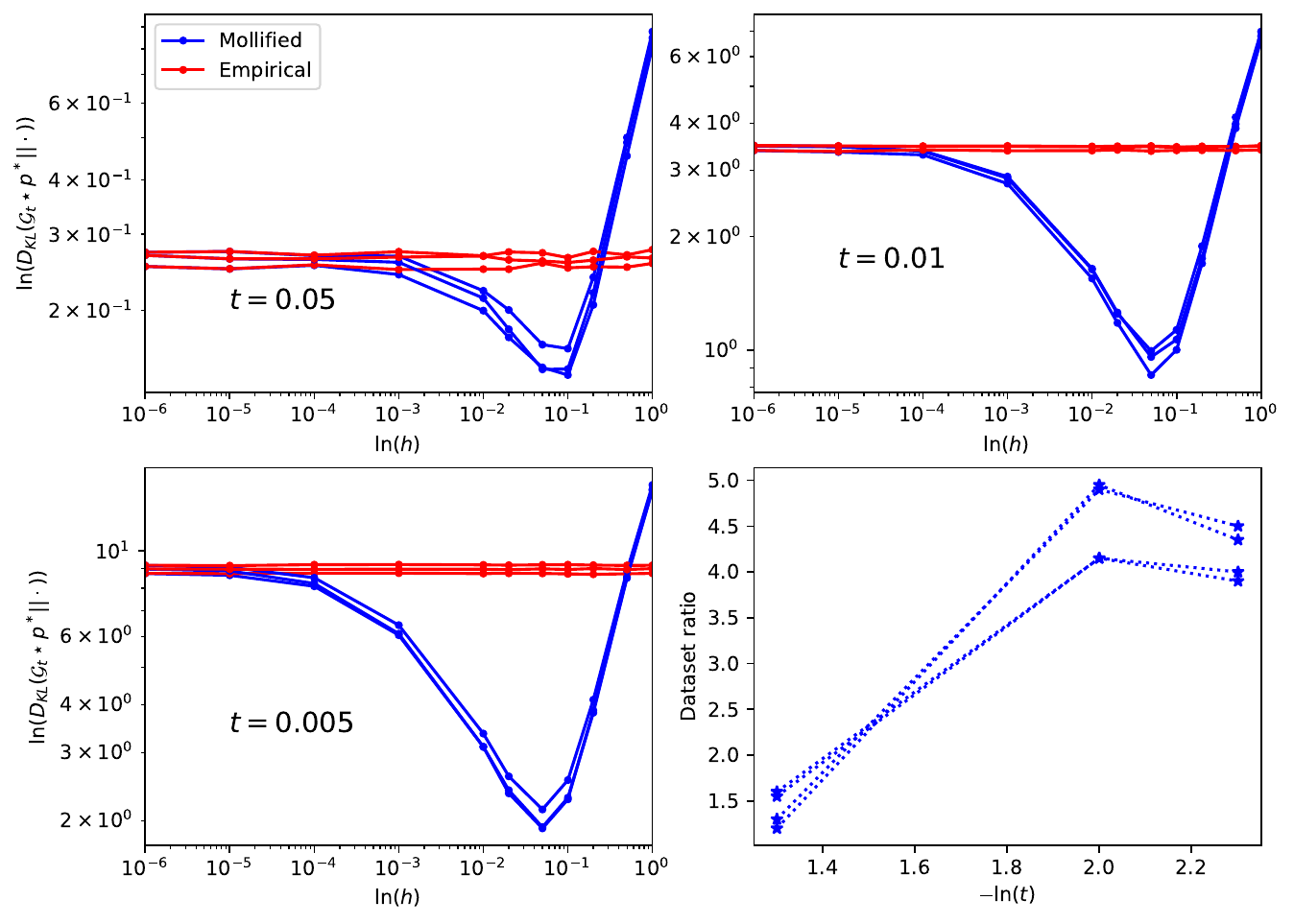}
\caption{1-3 figures: KL-divergence between $\mathcal{G}_{t_N}\star p_*$ and the  empirical measure
generated by following the score (red)  and the KL-divergence between  $\mathcal{G}_{t_N}\star p_*$ and the empirical measure generated by following the mollified score, varying $h$ (blue). 4th figure: Ratio $N_{\mathrm{eff}}/N$ at the lowest reported KL-divergence. $p_*$ is a uniform distribution over a 4-dimensional sphere with radius $1$, $N=100$ and $Q=10\ 000$ samples are used for the Monte-Carlo estimation of the density $p_{t}$.}\label{fig:sphere_d=10}
\end{figure}

In Figure \ref{fig:sphere_d=10}, we show again how the KL-divergence changes with respect to sampling time $t_N$ and the convolution bandwidth $h$, as well as the estimated $N_{\mathrm{eff}}$, showing that it is up to $4\times N$. %The magnitude of $N_{\mathrm{eff}}$ is similar to the Gaussian case, when $d=4$.

\subsection{Memorization}
\begin{figure}[h]
    \centering
\includegraphics[width=1.0\textwidth]{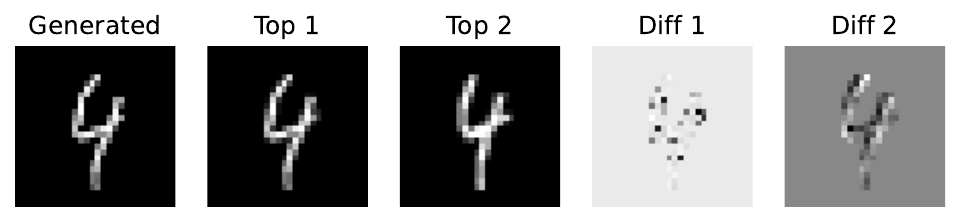}
\includegraphics[width=1.0\textwidth]{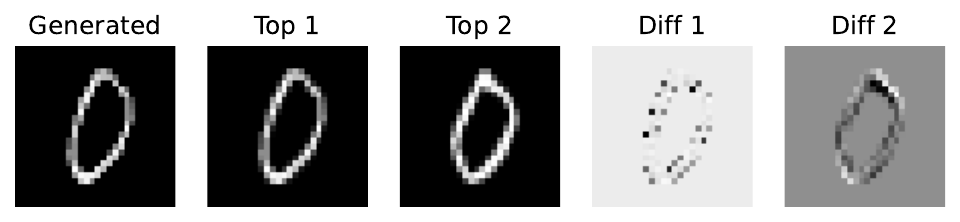}
\subfigure[Samples generated using the empirical score.]
{\includegraphics[width=1.0\textwidth]{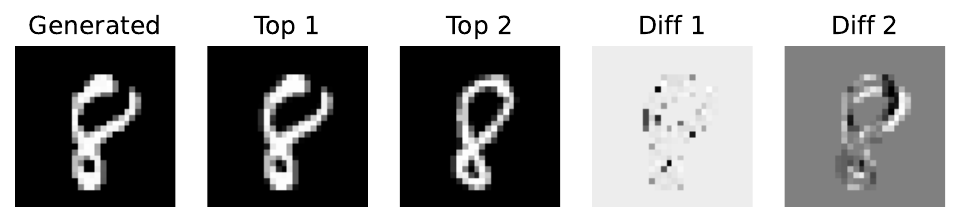}}
\includegraphics[width=1.0\textwidth]{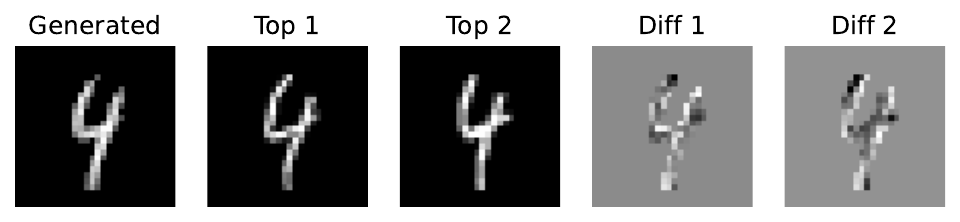}
\includegraphics[width=1.0\textwidth]{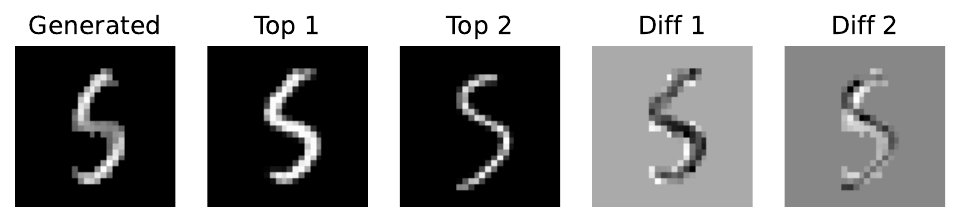}
\subfigure[Samples generated using the mollified score, $h=1.8$.]{\includegraphics[width=1.0\textwidth]{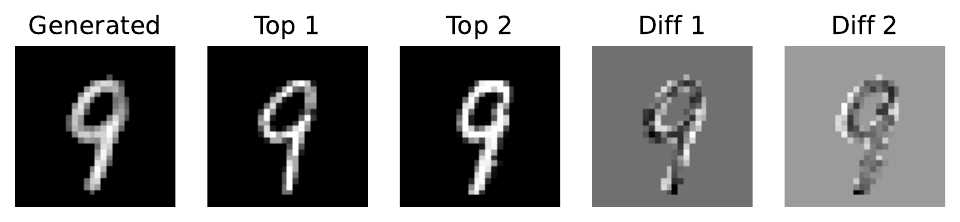}}
\caption{We set $t_N=10^{-3}$ and apply clamping to both samples, by setting all values below $0.25$ to $0$.}
\label{fig:mnist-generated-samples}
\end{figure}

In this experiment, we evaluate the effect of the mollified score on the memorization of the MNIST dataset. In Figure \ref{fig:mnist-generated-samples} we show generated samples from the MNIST dataset, using the empirical and mollified scores, as well as the two closest points in the training set to the generated sample, and their difference. It is shown that when following the empirical score, the diffusion model memorizes the dataset, as expected, whereas when following the mollified score, the generated samples appear to be some combination of elements on the training dataset, thus preventing pure memorization. 

\begin{figure}[h!]
    \centering
\includegraphics[width=0.7\textwidth]{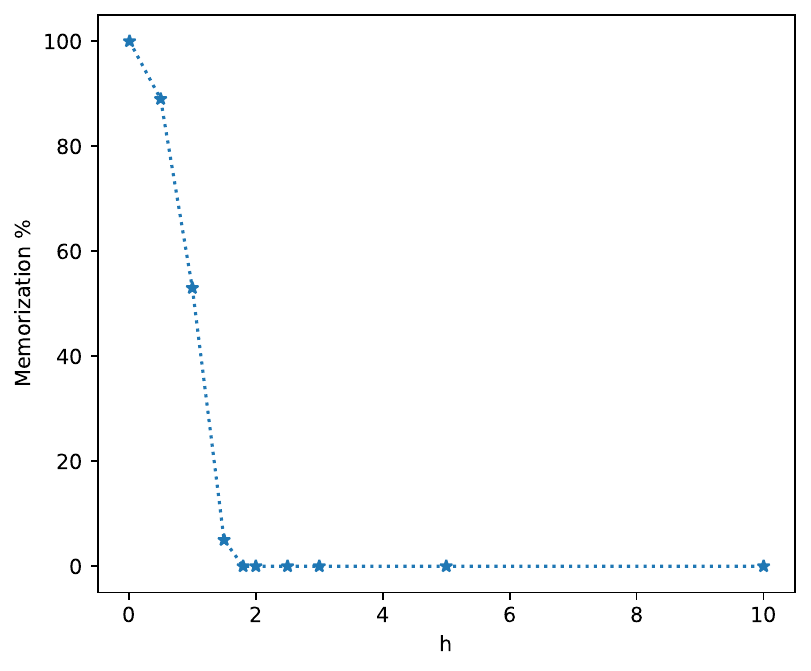}
\caption{Memorization ratio of $100$ generated samples, at $t_N=5\times 10^{-3}$.}
\label{fig:mnist-memorization}
\end{figure}

\begin{figure}[h!]
    \centering
\includegraphics[width=1.0\textwidth]{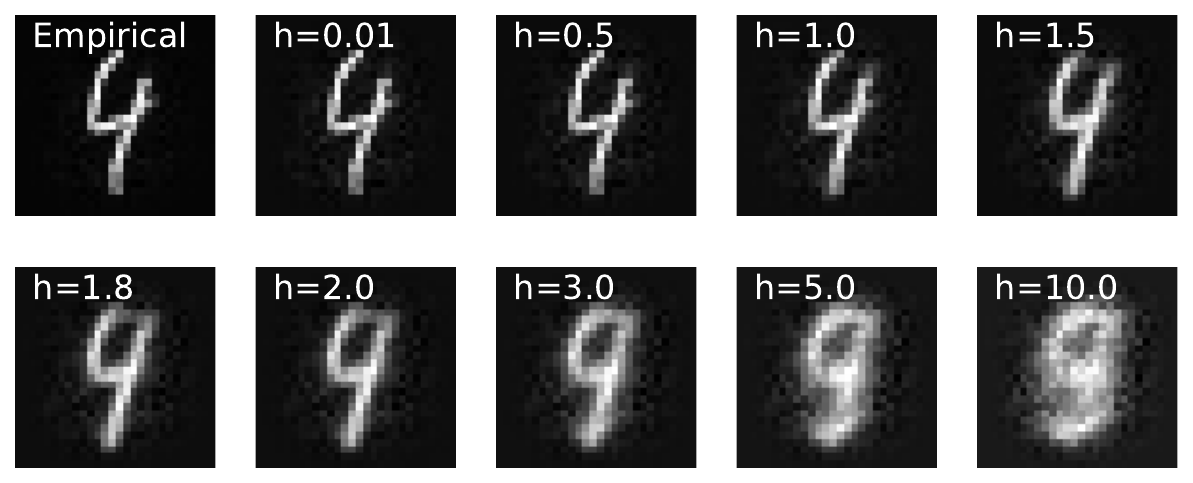}
\caption{Samples generated by starting at the same random initialization and following the corresponding score, without clamping values below $0.25$ to $0$. Sampling time $t_N=5 \times 10^{-3}$.}
\label{fig:examples-fixed-seed}
\end{figure}

In Figure \ref{fig:mnist-memorization}, we show the ratio of memorization while varying $h$. We use the memorization criteria as in \cite{yoon2023diffusion}, where a sample is considered memorized if $\frac{\lVert X-X_1 \rVert_2}{\lVert X-X_2 \rVert_2} <\frac{1}{3}$, where $X$ is the generated sample and $X_1$, $X_2$ are the first and second nearest neighbors in the training set. In Figure \ref{fig:examples-fixed-seed}, we show a generated sample starting with the same random initialization and varying $h$. It can be seen that as $h$ increases, the sample becomes more distinct from the training set, but also more noisy. At large $h$, the quality of the sample is significantly deteriorated. 

%\subsection{Scaling of Variance on Swiss Roll }

%\newpage

%%%%%%%%%%%%%%%%%%%%%%%%%%%%%%%%%%%%%%%%%%%%%%%%%%%%%%%%%%%%

%\appendix

%\section{Technical Appendices and Supplementary Material}
%Technical appendices with additional results, figures, graphs and proofs may be submitted with the paper submission before the full submission deadline (see above), or as a separate PDF in the ZIP file below before the supplementary material deadline. There is no page limit for the technical appendices.

%%%%%%%%%%%%%%%%%%%%%%%%%%%%%%%%%%%%%%%%%%%%%%%%%%%%%%%%%%%%

\end{document}